%% file: main.tex
\documentclass[10pt,twocolumn,letterpaper]{article}

\usepackage[pagenumbers]{cvpr} 
\usepackage{multirow} 
\usepackage{xcolor} 
\usepackage[table]{xcolor} 

\input{preamble}
\definecolor{cvprblue}{rgb}{0.21,0.49,0.74}
\usepackage[pagebackref,breaklinks,colorlinks,allcolors=cvprblue]{hyperref}

\def\paperID{20230} 
\def\confName{CVPR}
\def\confYear{2026}

\title{From Macro to Micro: Benchmarking Microscopic Spatial Intelligence on Molecules via Vision-Language Models}

\author{%
  Zongzhao Li$^{1,2,3}$\thanks{Equal contribution.} \quad
  Xiangzhe Kong$^{4,5}$\footnotemark[1] \quad
  Jiahui Su$^6$\quad
  Zongyang Ma$^{7}$ \quad 
  Mingze Li$^{1,2,3}$ \\
  Songyou Li$^{1,2,3}$ \quad
  Yuelin Zhang$^{1,2,3}$ \quad
  Yu Rong$^{8,9}$ \quad
  Tingyang Xu$^{8,9}$ \quad
  Deli Zhao$^{8,9}$ \quad
  Wenbing Huang$^{1,2,3}$\thanks{Corresponding author.}\\ 
  $^1$Gaoling School of Artificial Intelligence, Renmin University of China \\
  $^2$Beijing Key Laboratory of Research on Large Models and Intelligent Governance \\
  $^3$Engineering Research Center of Next-Generation Intelligent Search and Recommendation, MOE \\
  $^4$Dept. of Comp. Sci. \& Tech., Tsinghua University \\ 
  $^5$Institute for AI Industry Research (AIR), Tsinghua University \\ 
  $^6$SKL-ESPC \& SEPKL-AERM, College of Environmental Sciences and Engineering, Peking University \\ 
  $^7$MAIS, Institute of Automation, Chinese Academy of Sciences \\ $^8$DAMO Academy, Alibaba Group, Hangzhou,
China \ $^9$Hupan Lab, Hangzhou, China  \\
\tt\small lizongzhao2023@ruc\!.\!edu\!.\!cn,  jackie\_kxz@outlook\!.\!com,  hwenbing@126\!.\!com \\
}

\newcommand{\benchmark}{MiSI-Bench}

\begin{document}
\maketitle
\input{sec/0_abstract}    
\input{sec/1_intro}

\input{sec/2_related_work}
\input{sec/3_dataset}
\input{sec/4_Experiments}
\input{sec/5_Conclusion}
{
    \small
    \bibliographystyle{ieeenat_fullname}
    \bibliography{main}
}

\newpage
\appendix
\input{sec/suppl_A}

\input{sec/suppl_B}

\input{sec/suppl_C}

\end{document}

%% file: sec/0_abstract.tex
\begin{abstract}
This paper introduces the concept of Microscopic Spatial Intelligence (MiSI), the capability to perceive and reason about the spatial relationships of invisible microscopic entities, which is fundamental to scientific discovery. To assess the potential of Vision-Language Models (VLMs) in this domain, we propose a systematic benchmark framework \texttt{\benchmark}. This framework features over 163,000 question-answer pairs and 587,000 images derived from approximately 4,000 molecular structures, covering nine complementary tasks that evaluate abilities ranging from elementary spatial transformations to complex relational identifications. Experimental results reveal that current state-of-the-art VLMs perform significantly below human level on this benchmark. However, a fine-tuned 7B model demonstrates substantial potential, even surpassing humans in spatial transformation tasks, while its poor performance in scientifically-grounded tasks like hydrogen bond recognition underscores the necessity of integrating explicit domain knowledge for progress toward scientific AGI. The datasets are available at \url{https://huggingface.co/datasets/zongzhao/MiSI-bench}.
\end{abstract}

%% file: sec/1_intro.tex
\section{Introduction}
\label{sec:intro}

\begin{figure*}[]
    \centering    \includegraphics[width=\linewidth]{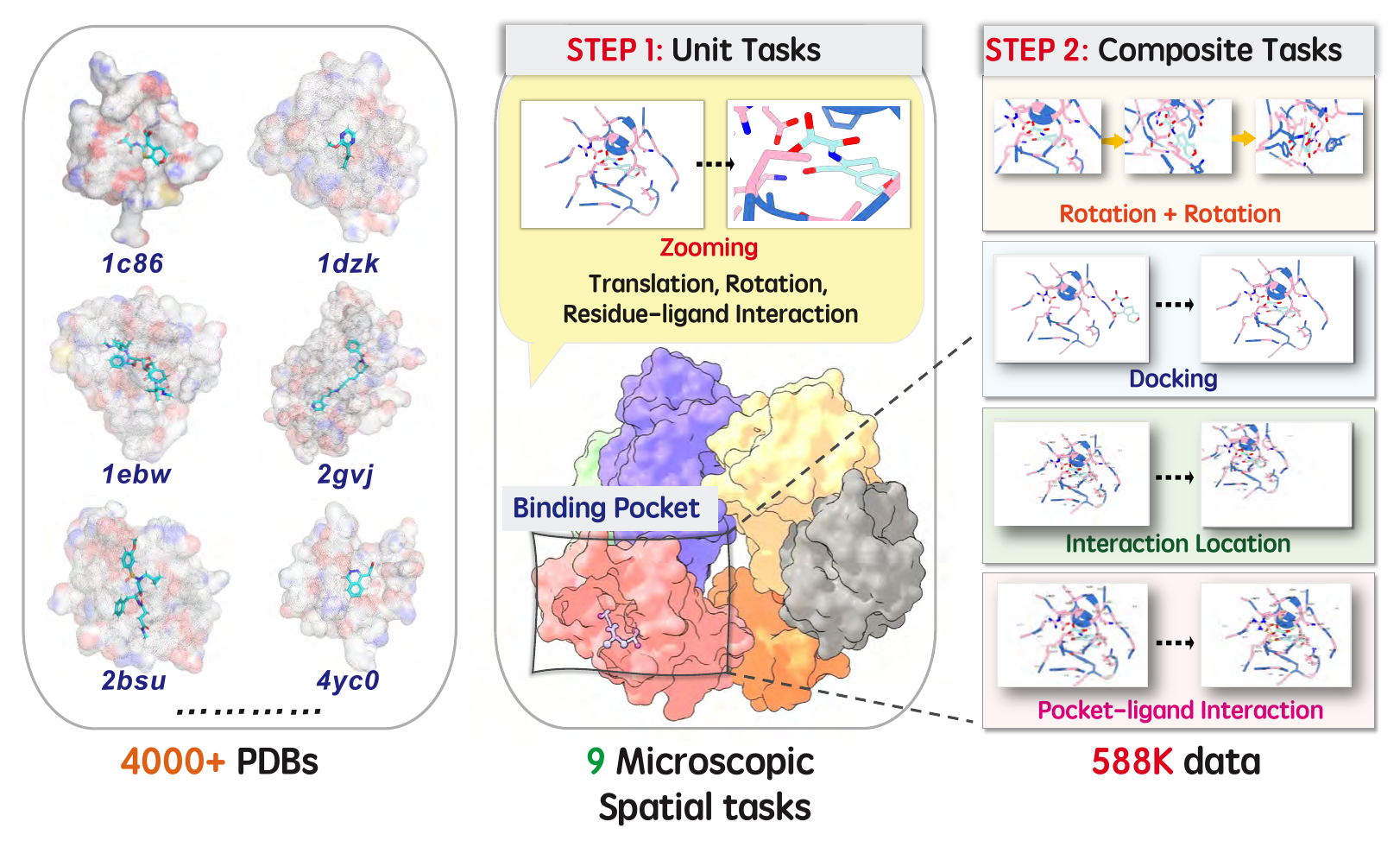}
    \vspace{-15pt}
    \caption{Overview of our \texttt{\benchmark}. Our dataset is derived from around 4,000 PDBs and comprises 9 distinct tasks.}
    \label{fig:model}
    \vspace{-10pt}
\end{figure*}

Spatial intelligence~\cite{chen2024spatialvlm, yang2025thinking}, a critical component of advanced artificial intelligence, empowers systems to perceive, interpret, and interact with the three-dimensional world. Such capability necessitates a profound comprehension of geometries, spacial relationships, and even fundamental physical rules. Contemporary efforts address this by utilizing Vision-Language Models (VLMs)~\citep{bai2025qwen25vl, li2024llava-next, li2024llava-onevision}, which jointly process visual and textual data to learn spatial properties and relationships from complex scenes~\citep{yin2025spatial, zhang2025flatland, feng2025onethinker, wu2025reinforcing}. This reasoning capacity regarding object layouts, occlusions, and perspectives establishes a vital foundation for embodied interaction in real-world environments~\citep{zhao2025embodied}.

Yet, beyond this visible macroscopic realm lies the microscopic world, composed of invisible particles (e.g., atoms and molecules) that constitute matter~\citep{eidelman2004review}, where spatial reasoning takes on a profoundly different form. In molecular sciences, experts routinely visualize and manipulate microscopic entities such as proteins and drugs using software tools (\emph{e.g.}, PyMOL~\citep{delano2002pymol}, ChimeraX~\citep{pettersen2021ucsf}) to explore geometric complementarity, analyze interactions, and design new functional molecules. This process relies on a specialized form of spatial reasoning: the ability to reconstruct three-dimensional structures from two-dimensional projections and to infer physical relationships (\emph{e.g.}, hydrogen bonds). In this paper, we refer to this capability as \textbf{Microscopic Spatial Intelligence (MiSI)}, the cognitive foundation underlying human expertise and discovery in scientific fields such as structural biology, drug discovery, and material design.

The success of Large Language Models (LLMs) in general-purpose tasks has spurred their exploration in scientific discovery~\citep{boiko2023autonomous, m2024augmenting, swanson2025virtual}, motivating the use of VLMs to analyze scientific data. VLMs are uniquely suited for this role, as they can process both visual and textual modalities within a unified architecture. Unlike conventional domain-specific systems, VLMs can perceive structural patterns while grounding their interpretations in scientific concepts. This cross-modal capability enables a more human-like, context-aware reasoning about molecular structures by seamlessly linking them with textual semantics. However, shifting from human-scale daily objects to atom-level invisible entities, MiSI requires exceptional scientific expertise to perceive \textit{spatial transformations} and reason over \textit{relational identifications} such as atomic interactions. It remains unclear whether the VLMs are ready for tackling the challenges in the microscopic scientific fields.

To bridge the gap, we propose \texttt{\benchmark}, a systematical framework for training and evaluating microscopic spatial intelligence in VLMs. As illustrated in~\cref{fig:model}, \texttt{\benchmark} contains 163,514 question-answer pairs and 587,975 images over a diverse set of 9 complementary tasks, constructed from around 4,000 molecular structures~\citep{wang2005pdbbind}. We visualize these three-dimensional microscopic objects as two-dimensional orthographic projections, just like how human experts interpret them. We then disentangle the intelligence for \textit{spatial transformations} and \textit{relational identifications} into four elementary operations: translation, rotation, zooming, interaction. Subsequently, we design four unit tasks to evaluate these fundamental abilities independently, and further design five composite tasks which integrate multiple elementary operations to test the models' high-order reasoning ability.

Experimental results show that current advanced VLMs (e.g., o3~\citep{o3}, Claude Sonnet4.5~\citep{Claude-4.5-Sonnet}) perform well below human level on our benchmark. While human evaluators excel in some tasks, they struggle with continuous spatial transformation and 3D reconstruction. Remarkably, after SFT on our dataset, a 7B model outperforms all leading VLMs and even surpasses humans in spatial transformation tasks, revealing VLMs' untapped potential for spatial reasoning. However, its poor performance in biologically-grounded tasks like hydrogen bond recognition highlights the need for injecting explicit scientific knowledge during pre-training to progress toward AGI.

%% file: sec/2_related_work.tex
\section{Related Work}
\label{sec:related_work}

\paragraph{Macroscopic Spatial Intelligence}
In recent years, researchers have developed a variety of datasets and benchmarks with distinct focuses to evaluate the spatial intelligence of VLMs~\citep{zuo2025towards, ray2024sat, fu2024blink}. For instance, VIS-Bench~\citep{yang2025thinking} and MuriBench~\citep{wang2024muirbench} emphasize the model’s ability to associate and reason across video/multi-images; LEGO-Puzzles~\citep{tang2025lego} examines multi-step spatial reasoning in a synthetic block-building environment. Therefore, we propose \texttt{\benchmark} to draw attention to this direction and to establish a reliable benchmark for evaluating models’ micro-level spatial intelligence.
This task is uniquely challenging, as understanding microscopic entities demands exceptional expertise in both spatial transformations and relational reasoning.

\paragraph{Three-Dimensional Molecular Understanding}~Conventional modeling of 3D molecules usually relies on Cartesian coordinates as model inputs. Starting from physical force fields~\citep{alford2017rosetta, schymkowitz2005foldx}, models have evolved from 3D convolutional neural networks~\citep{pinheiro2024structure} to equivariant graph neural networks~\citep{kong2023conditional, tholke2022equivariant} and transformers~\citep{kong2025unimomo, huang2025equivariant} with the rise of geometric deep learning~\citep{bronstein2017geometric}. However, these approaches primarily operate within geometric coordinate space only, while MLLMs offer a complementary, human-like perspective which can learn to reason about three-dimensional molecular geometry through visual abstractions and natural language grounding, unlocking a new mode of molecular understanding that bridges microscopic visual perception and textual knowledge.


%% file: sec/3_dataset.tex
\section{Definition of Major Concepts}
\label{sec:msi}

\begin{figure*}[ht!]
    \centering    \includegraphics[width=1.0\linewidth]{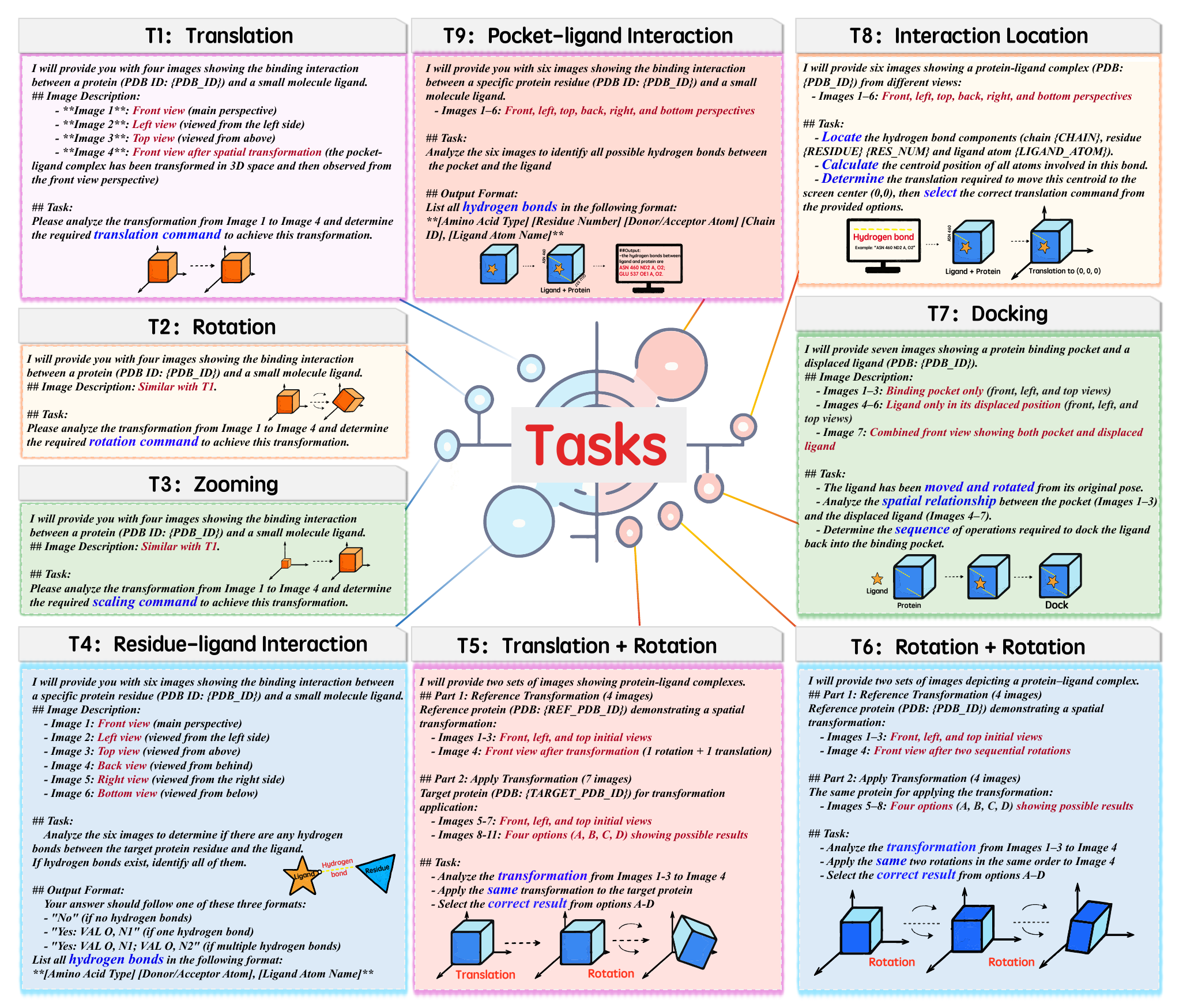}
    \vspace{-15pt}
    \caption{A brief demonstration of the \texttt{\benchmark} dataset. The examples have been simplified for clarity; for complete examples, please refer to Appendix C.}
    \label{fig:dataset}
\end{figure*}

\begin{figure}[h!]
    \centering
    \includegraphics[width=\columnwidth]{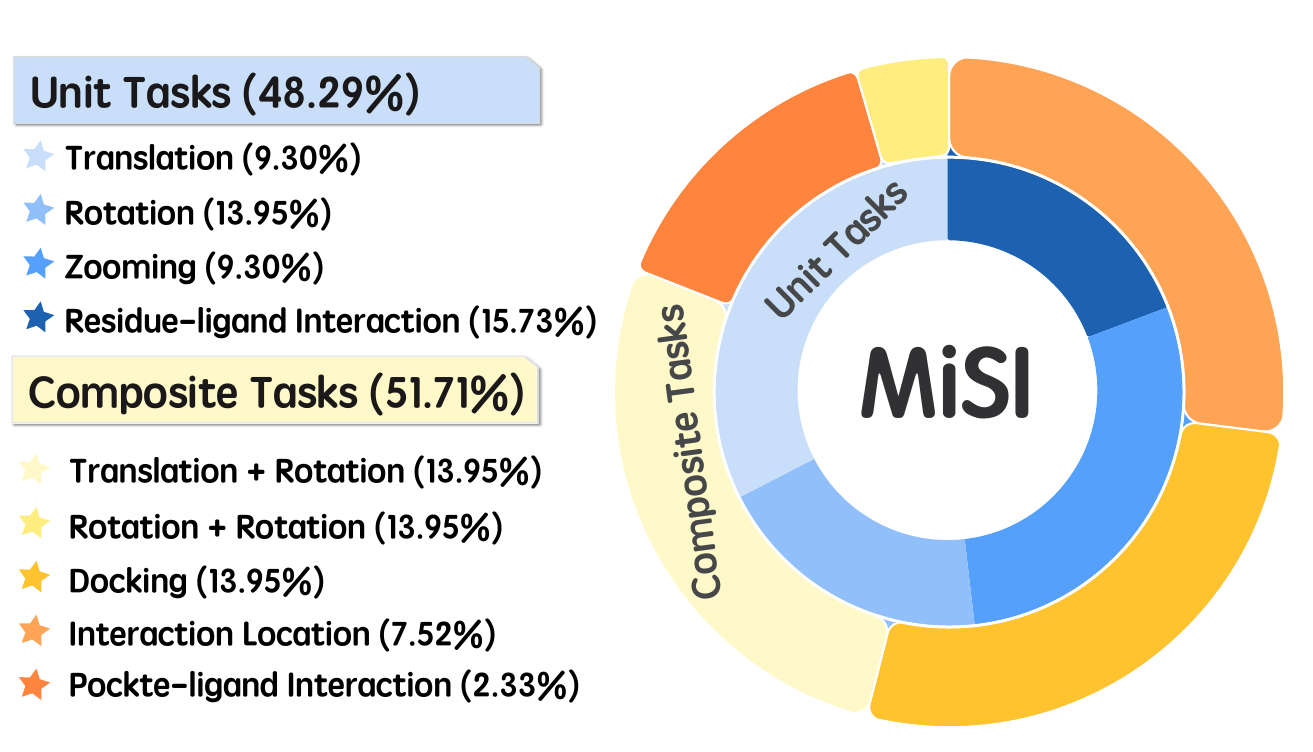}
    \caption{The statistics for all tasks.}
    \label{fig:statics}
    \vspace{-15pt}
\end{figure}

To explore whether current VLMs possess the ability to understand 3D molecular structures, which we term \textbf{MiSI} as above, we begin by defining the representations adopted for molecular structures, and then introduce the fundamental perceptual expertise humans rely on to comprehend the micro world. These elements together serve as the preliminaries for our benchmark design. 

\paragraph{Study Scope}
Our physical world is organized across multiple hierarchical levels, from macroscopic organisms to organs, cells, and to molecules such as proteins, and DNAs~\citep{eidelman2004review}. Thanks to the continuous scientific exploration, people now understand the phenomena observed in the macroscopic world are the results of microscopic particles. And advances in imaging technologies, such as cryo-electron microscopy, further allow us to visualize these particles at near-atomic resolution~\citep{bai2015cryo}. In this work, we shift the focus from the macroscopic world to its microscopic foundation, investigating how VLMs perceive and reason about 3D molecular structures composed of atoms.

\paragraph{Orthographic Projection of Molecules}
Throughout human history, people have sought to represent the three-dimensional world on two-dimensional media. One classical approach employs perpendicular rays of light to generate \textit{orthographic projections}, and typically canonical views, such as the front, top, and left views, are employed to reconstruct the full 3D structure of an object~\citep{carlbom1978planar}. Following this convention, we adopt orthographic views as 2D representations of microscopic 3D molecular structures.

\paragraph{Taxonomy of Human Expertise}
Understanding microscopi molecules requires both spatial reasoning and domain expertise. Experts rely on fundamental \textit{spatial transformation} abilities, such as translation, rotation, and zooming, to establish a more complete panorama of molecular structures~\citep{cherezov2007high, kong2025peptide}. Beyond geometric manipulation, they use domain knowledge to identify interaction patterns such as hydrogen bonds, which reveal the underlying physical principles of molecular organization~\citep{anderson2003process}. We refer to this process as \textit{relational identification}. In this work, we summarize the expert skills with the above-mentioned four elementary microspace operations, namely \texttt{translation}, \texttt{rotation}, \texttt{zooming}, and \texttt{interaction}, then design unit tasks to independently evaluate each capability. We further introduce combinatorial tasks that require integrating multiple operations, enabling a more comprehensive assessment of VLMs in microspace understanding.


\section{\texttt{\benchmark}}
\label{sec:benchmark}
We construct our \texttt{\benchmark} for evaluating VLMs using the refined PDBbind dataset~\citep{wang2005pdbbind}, a widely adopted benchmark for structure-based drug discovery~\citep{kong2024generalist, gao2023drugclip, townshend2021atomd}. Each entity in PDBbind dataset corresponds to a complex composed of a protein and a ligand. We visualize all complexes in ChimeraX~\citep{pettersen2021ucsf} to generate orthographic projections images as model inputs. After removing complexes with visualization issues, the final dataset contains 3,503 protein–ligand complexes for Supervised Fine-Tuning (SFT) and 490 for testing, all with experimentally solved crystal structures. The benchmark encompasses nine tasks, including four unit tasks involving single elementary operations and five composite tasks involving combinations of multiple operations. For each task, the QA pairs are generated using fixed templates. The problem templates and a brief illustration of all tasks are shown in ~\cref{fig:dataset}. The overall pipeline for constructing \texttt{\benchmark} is detailed in supplementary materials. Our benchmark contains a total of 150,597 Question Answering (QA) pairs for training and 12,917 for testing, summing up to 538,015 and 49,960 images in the train and the test set, respectively. The statistics for all tasks are presented in ~\cref{fig:statics}. Two formats of questions are used to evaluate model performance.

\textit{Cloze Questions} require the model to complete partially specified instructions by filling in missing actions or parameters. These tasks assess the ability of the models to identify the correct operations with precise attributes (e.g., axes, distances, or angles).

\textit{Multiple-Choice Questions} present several candidate options, among which the model should identify the correct answer while rejecting distracting decoys. These tasks evaluate whether the models have discriminative understanding of spatial configurations and their ability to reason about the consequences of microspace operations.

\subsection{Unit Task}
\label{sec:unit_task}

We first establish four unit tasks to evaluate the spatial understanding of elementary microspace operations, where each task isolates one essential ability involved in manipulating or interpreting microscopic 3D molecular structures. For translation, rotation, and zooming tasks, the three orthographic projections (\emph{i.e.}, the top, the front, and the left side views) of the initial complex and the front view of the complex after the operation are given to the models. For residue-ligand interaction task, since overlapping atom names might interfere with the performance, we give all six orthographic projections to the models. 

\paragraph{Translation~\textnormal{(\textit{Cloze})}} In this task, the molecular complex is translated along one of the axes parallel to the visualization plane (\emph{i.e.}, the $x$ or $y$ axis) by a random distance between $-4$ and $4$ angstrom (\AA). The model must infer both the direction and the magnitude of motion, completing a prompt of the form:
\texttt{move \underline{ x } \underline{ 3 }}.
To avoid being too harsh on numerical precisions, the translation range is discretized into 1.0~\AA~bins.
For SFT, two samples per complex are generated along each axis, yielding 14,012 training samples, and one per axis for evaluation, totaling 980 test samples.

\paragraph{Rotation~\textnormal{(\textit{Cloze})}} The complex is rotated along one of the three coordinate axes ($x$, $y$, or $z$) by a random angle uniformly drawn from $[-90^\circ, 90^\circ]$. Models must determine both the rotation axis and the degree of rotation, filling in the prompt: \texttt{roll \underline{ x } \underline{ 15 }}. Rotation angles are discretized into $15^\circ$ bins. Each complex results in two samples per axis for SFT (21,018 training samples) and one per axis for testing (1,470 samples).

\paragraph{Zooming~\textnormal{(\textit{Cloze})}} To simulate zooming operations, the complex is moved along the axis perpendicular to the visualization plane (\emph{i.e.}, the $z$ axis) by a random depth between 40 and 60~\AA. This range corresponds to the magnification levels most suitable for visualizing the pocket–ligand interactions near the center of the view (See the distribution figure in Appendix A for details). The model fills in prompts like: \texttt{move z \underline{ 50 }}, where depth values are discretized into 1.0~\AA\ bins. Four samples per complex are created for SFT (14,012 training samples) and two per complex for testing (980 samples).

\paragraph{Residue-Ligand Interaction~\textnormal{(\textit{Cloze})}} Given a residue and the ligand, models must first identify whether the residue interacts with the ligand (\texttt{\underline{Yes}} or \texttt{\underline{No}}), and then output all atom pairs participating in the interaction: \texttt{\underline{ARG NH2, O22; ARG N, O23}}.
For this proof-of-concept benchmark, we focus on hydrogen bonds as interactions of interest, with detailed geometric configurations provided in the Appendix A. The dataset includes 11,572 positive and 12,125 negative samples for SFT, and 1,499 positive and 1,603 negative samples for evaluation.

\subsection{Composite Tasks}
\label{sec:composite_task}
We further design five composite tasks that require models to understand and reason on multiple microspace operations, the capability of which are commonly required for human experts during molecular structural analysis.

\subsubsection{Spatial Transformation Reasoning}

This category evaluates the ability of the models to reason about sequential spatial transformations and generalize them across different molecular complexes. Denote two complexes as $c_1$ and $c_2$, and two spatial transformations as $f_1$ and $f_2$. The models are given the three orthographic projections of both $c_1$ and $c_2$, along with the front view of $f_2(f_1(c_1))$, which is the result of applying $f_1$ followed by $f_2$ to $c_1$. The task is to identify the correct front view of $f_2(f_1(c_2))$ among four candidate images, where the other three are decoys. The decoys are constructed by exerting perturbation on the ground-truth transformations through one of three schemes: 1) Altering the magnitude (translation distance or rotation angle) of both $f_1$ and $f_2$; 2) Flipping the sign of $f_1$ (e.g., clockwise to counterclockwise) and adjusting the magnitude of $f_2$; 3) Changing the axis of $f_1$ and modifying the magnitude of $f_2$.

\paragraph{Translation-Rotation Movement~\textnormal{(\textit{Multiple-Choice})}} In this task, $f_1$ is sampled from translational operations and $f_2$ from rotational operations. For each complex in the dataset, we pair it with another random complex and construct six and three distinct transformation combinations for SFT and testing, respectively, yielding a total of 21,018 and 1,470 questions for SFT and testing.

\paragraph{Rotation-Rotation Movement} \textit{(Multiple-Choice)} Both $f_1$ and $f_2$ are sampled from rotational operations, constrained to different axes to prevent trivial correlations. The scale matches the previous task, with 21,018 questions for SFT and 1,470 for testing.

\subsubsection{Local Relational Reasoning}

This category evaluates whether the models are capable of interpreting fine-grained, domain-specific spatial relations within molecular complexes, such as hydrogen bonds between atom pairs, and then reasoning about how to manipulate the visualization to highlight specific interactions.

\paragraph{Interaction Location~\textnormal{(\textit{Multiple-Choice})}} The model is provided with six orthographic projections of a molecular complex, along with a specified atom pair representing a hydrogen bond (e.g., \texttt{ARG 45 NH2 A, O1B}, denoting the interaction between the \texttt{NH2} atom of residue \texttt{ARG45} on chain \texttt{A} and the \texttt{O1B} atom on the ligand). The objective is to distinguish the correct transformation that repositions the corresponding interaction to the center of the visualization from three other decoys. The decoy transformations are generated by perturbing the sign and magnitude of the ground-truth translation parameters.

\subsubsection{Global Relational Reasoning}

This series of tasks considers the overall spatial relations of the entire complex instead of single localized ones, which is inherently more difficult than previous tasks, as they test the ability of the models to reason high-order combinations of spatial operations.

\paragraph{Ligand Docking~\textnormal{(\textit{Cloze})}} This task emulates the molecular docking process to evaluate whether the model can infer the complementary binding configuration and corresponding geometric transformations. The model is provided with three orthographic views of the ligand alone, the pocket alone, and one undocked complex view obtained by translating and rotating the ligand away from its pocket. Rotation angles are uniformly sampled from $[-90^\circ, 90^\circ]$, while translation distances are adaptively determined for each complex to minimize spatial overlap between the displaced ligand and pocket (Specific details can be found in Appendix A). The model must predict the sequence of transformations required to recover the native docking conformation, such as \texttt{roll y 45, move x -12}. Each complex generates six training samples (21,018 in total) and three test samples (1,470 in total).

\paragraph{Pocket-Ligand Interaction~\textnormal{(\textit{Cloze})}} This task extends the Residue–Ligand Interaction task to the entire binding pocket, requiring the model to reason about global intermolecular contact patterns. Given six orthographic projections of the full complex, the models are required to output all hydrogen-bond interactions between the ligand and the pocket in a structured format like \texttt{ARG 221 NH2 A, O22; ARG 221 N A, O23; ARG 221 NE A, O22}.
Each interaction is expressed as a tuple specifying the residue type, residue index, interacting atom in the residue, chain identity, and interacting atom in the ligand, with semicolons separating multiple entries.

%% file: sec/4_Experiments.tex
\section{Experiments}
\input{Table/Main_results}
We first introduce the experimental setup in~\cref{sec:setup}, and then report the main results of all compared models on our benchmark in~\cref{sec:results}. Finally, we conduct factor analysis and case study in~\cref{sec:analysis}.

\subsection{Setup}
\label{sec:setup}
\paragraph{Benchmark Subsets}~Due to the expensive cost of closed-source models (especially reasoning models), we create \texttt{\benchmark (tiny)} by randomly sampling 50 question-answer pairs from each task in our dataset. This tiny subset will be used for evaluating the performance of all closed-source models and open-source mixture-of-experts (MoE) models, ensuring an intuitive and fair comparison. All models are evaluated under few-shot settings to provide them with necessary scientific prior knowledge.

\paragraph{Metrics}~For \textit{Multiple-Choice Questions} and Zooming task, we follow the convention to adopt Accuracy (ACC) as the main metric~\citep{yue2024mmmu, fu2025video}, which is calculated as the proportaion of answers exactly matching the ground truth. For \textit{Cloze Questions}, where answers might involve continuous numerical values and multiple entries, we employ a weighted composite score, as inspired by previous literature~\citep{everingham2010pascal, lin2014microsoft}, to reflect the degree of correctness beyond exact matching. For tasks involving spatial transformations (\emph{i.e.}, \texttt{move} and \texttt{roll}), when the model predicts the correct axis, we will further assign scores based on the predicted values. Specifically, the score is determined by the normalized absolute error between the predicted ($\hat{d}$) and ground-truth ($d$) magnitudes (\emph{i.e.}, distances and angles), $|\hat{d} - d|$. For composite tasks involving multiple transformations, such as Ligand Docking, each component operation (\emph{e.g.}, \texttt{move} and \texttt{roll}) contributes equally to the total score, summing up to $1.0$. In this task, since the axis for the \texttt{move} operation is fixed to $x$, scores are assigned only when the sign of the predicted ($\hat{d}$) matches the sign of the ground-truth ($d$). For Residue–Ligand Interaction and Pocket–Ligand Interaction tasks, we compute the ratio of correctly predicted interactions among all provided outputs. To penalize cheating behaviors where models output all the correct interactions along with hallucinations of irrelevant ones, such cases are assigned with a score of $0.5$. Furthermore, if the number of hydrogen bonds in the model's response exceeds twice the number in the ground-truth, we consider the model to be attempting to score through exhaustive enumeration, in which case the model receives a score of $0$. Furthermore, we also provide the results for exact matching for \textit{Cloze Questions} in Appendix B.


\paragraph{Benchmark Models}~We comprehensively evaluate ten VLMs spanning four major model families, including nine closed-source and one open-source representatives. From Open AI, we include GPT-5-mini (w/ mixed modes of reasoning)~\citep{gpt5}, o4-mini (w/ reasoning)~\citep{o3}, o3 (w/ reasoning)~\citep{o3}, and GPT-4.1 (w/o reasoning)~\citep{GPT-4.1}. From Anthropic’s Claude series, we test Claude 4.5 Sonnet (w/ reasoning)~\citep{Claude-4.5-Sonnet}, Claude 4 Opus (w/ reasoning)~\citep{Claude-4-Opus}, and Claude 3.5 Sonnet (w/o reasoning)~\citep{Claude-3.5-Sonnet}. From Google’s Gemini series, we include Gemini-2.5-pro (w/ reasoning)~\citep{Gemini-2.5-and-2.5-Pro} and Gemini-2.5-flash-lite (w/ reasoning)~\citep{comanici2025gemini}. For open-source models, our preliminary experiments show that most models below 32B parameters achieve very low performance on \texttt{\benchmark}. Therefore, we select the strong Qwen3-vl-235b-a22b-thinking (w/ reasoning)~\citep{Qwen3-VL} from larger open-source models as a representative baseline. Furthermore, we propose Qwen2.5VL-7B-SFT, which is finetuned on the training split of our benchmark to investigate how to better trigger the \textit{Microscopic Spatial Intelligence} in VLMs.

\paragraph{Human-Level Performance}~We estimate the performance of humans by recruiting PhD candidates in biology to complete the questions for residue-ligand interaction, ligand docking, and pocket-ligand interaction, which requires more domain expertise than the other tasks. For the rest of tasks, we employ PhD candidates in broader field of science, technology, engineering, and mathematics to propose answers. Each participant is required to answer questions in \texttt{\benchmark (tiny)} independently, whose responses are later assessed with the same metrics as the models.

\subsection{Main Results}
\label{sec:results}

\textbf{Human Level Performance.}
The evaluation results are shown in ~\cref{tab:main_results}. Human evaluators perform well in most unit tasks, demonstrating strong 3D spatial modeling abilities and the potential to integrate biological knowledge with spatial reasoning for basic interactive tasks. However, their performance decline significantly in complex spatial reasoning tasks. They manage small-angle rotations by tracking key atomic changes, while large-scale rotations—requiring maintained spatial continuity and multi-atom tracking—increase cognitive load and impaired axis and angle judgment. Zooming tasks prove even more challenging, as their judgments rely on overall intuition regarding boundary shifts and atomic density changes, lacking clear reference points and resulting in greater estimation errors.


In composite tasks, consecutive spatial operations (e.g., Trans-Rot., Rot-Rot.) lead to error accumulation and frequent reference frame shifts, significantly degrading human performance. Such tasks impose high demands on working memory and the stability of spatial mental simulation, forming a bottleneck in human performance. In Docking task, performance is the poorest due to the need for both sequential spatial transformations and biological knowledge to determine hydrogen bond formation and optimal docking positions. For Poc-Lig Inter. task, the main challenge lies in integrating multiple 2D views to reconstruct the 3D conformation of occluded residues before identifying hydrogen bonds, making it highly demanding. In contrast, Inter Location. task are simpler: they involve no rotational operations—only distance perception—and provide hydrogen bond information upfront, eliminating the need for specialized knowledge.


\textbf{Advancing VLMs Performance.}
As evidenced by the results in the table, all advanced models perform sub-optimally across various tasks in \texttt{\benchmark (tiny)}. Overall, the models exhibit better performance in distance-related tasks compared to rotation-related ones. For instance, in tasks such as "Translation" versus "Rotation", and "Interaction Location" versus "Rotation–Rotation Movement," the models consistently achieve higher scores in the former. This may stem from the fact that most current VLMs are primarily trained on two-dimensional data, making distance—as a two-dimensional attribute—more readily adaptable for the models. Furthermore, in the "Residue–Ligand Interaction–Pos" and "Pocket–Ligand Interaction" tasks, the performance gap between the models and human-level performance is most pronounced, highlighting their still-insufficient knowledge reserves in specialized domains such as biology. In contrast, in the "Residue–Ligand Interaction–Neg" task, most models perform relatively well, likely because the greater spatial distance between residues and ligands in negative samples allow the models to make correct judgments based solely on spatial proximity.


\textbf{SFT Model Performance.}
Experimental results demonstrate that after fine-tuning on the \texttt{\benchmark} dataset, model performance improves significantly, surpassing mainstream VLMs across all tasks and exceeding human-level performance in complex spatial tasks such as Rotation. Notably, in the Rot–Rot. task, where human performance approaches random guessing, the model maintains approximately 90\% accuracy, indicating that advanced VLMs possess potential for 3D spatial cognition. Previous underperformance of advancing VLMs may have stemmed from domain adaptation barriers: although models have generic spatial understanding, they lack visual priors for specialized structures such as proteins, hindering knowledge transfer. Appropriate fine-tuning can establish cross-domain mappings and unlock their spatial reasoning capabilities. However, in tasks such as Res/Poc–Lig Inter., which rely on domain-specific knowledge, models still lag behind humans, suggesting that the absence of domain priors in foundational training remains a bottleneck. Future work should focus on further exploring the spatial potential of models and investigating how to effectively integrate explicit knowledge from scientific fields such as structural biology.

\subsection{Analysis}
\label{sec:analysis}
\begin{figure}[]
    \centering
    \includegraphics[width=\columnwidth]{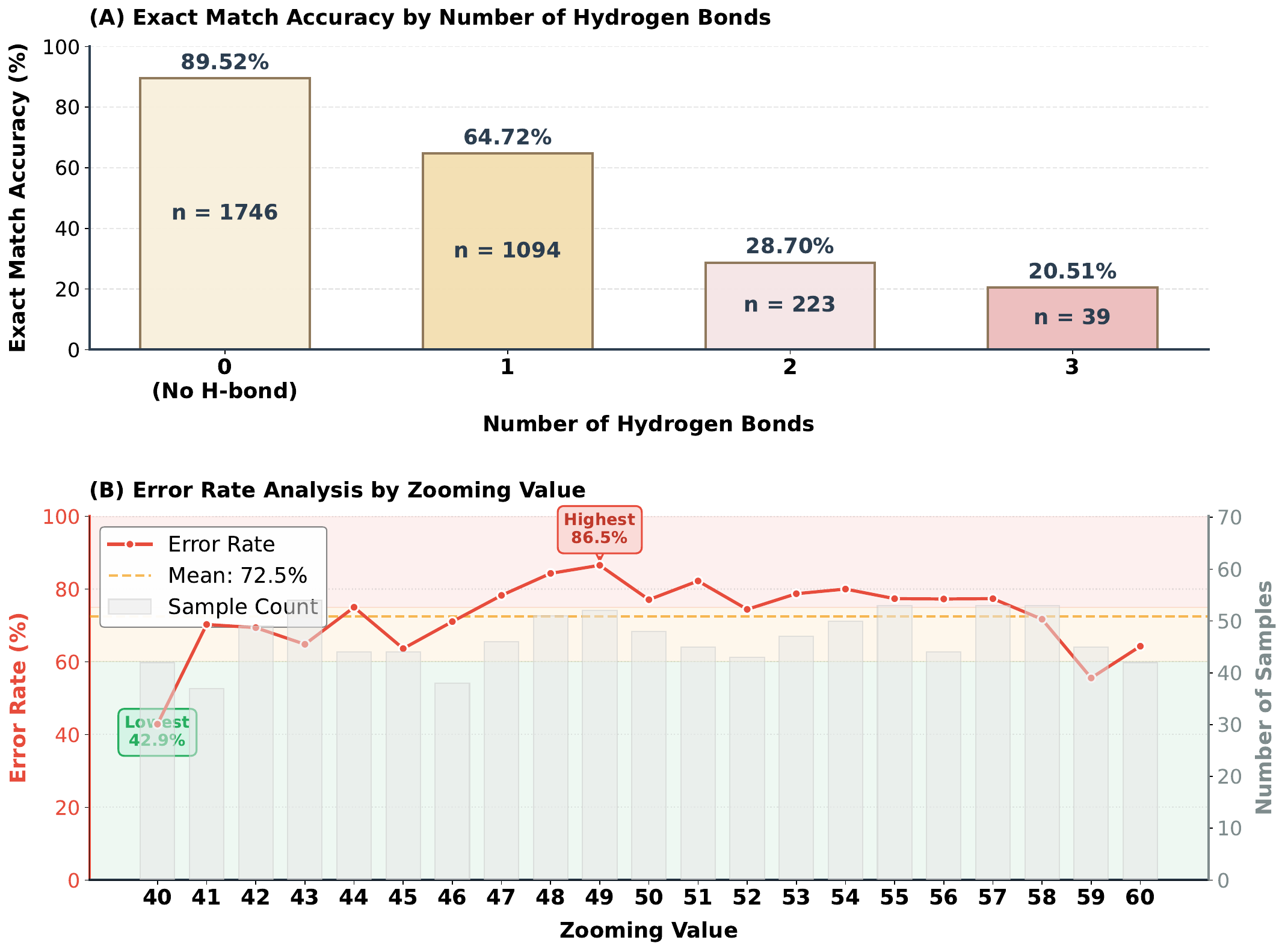}
    \vspace{-15pt}
    \caption{Factor Analysis of \texttt{\benchmark}.}
    \label{fig:analysis}
    \vspace{-15pt}
\end{figure}

\textbf{Factor Analysis.} 
In this section, we conduct a detailed analysis of the suboptimal performance exhibited by the SFT model on the Res-Lig Inter Pos. and Zooming tasks. ~\cref{fig:analysis}(a) and (b) present the prediction accuracy of the model across different statistical intervals for these two tasks. As shown in ~\cref{fig:analysis}(a), the prediction accuracy decreases sharply as the number of hydrogen bonds increases, indicating that the model struggles to identify all hydrogen bonds in scenarios with complex hydrogen-bonding interactions. In ~\cref{fig:analysis}(b), the prediction error rate curve exhibits an initial increase followed by a decrease. We hypothesize that this pattern may stem from the model’s failure to generalize uniformly across the entire scale space. The observed peak likely corresponds to a visually critical scale in molecular structures, where discriminative structural information is minimal, thereby reducing the parsing efficiency of the model’s attention mechanism.

\begin{figure}[h]
    \centering
    \includegraphics[width=\columnwidth]{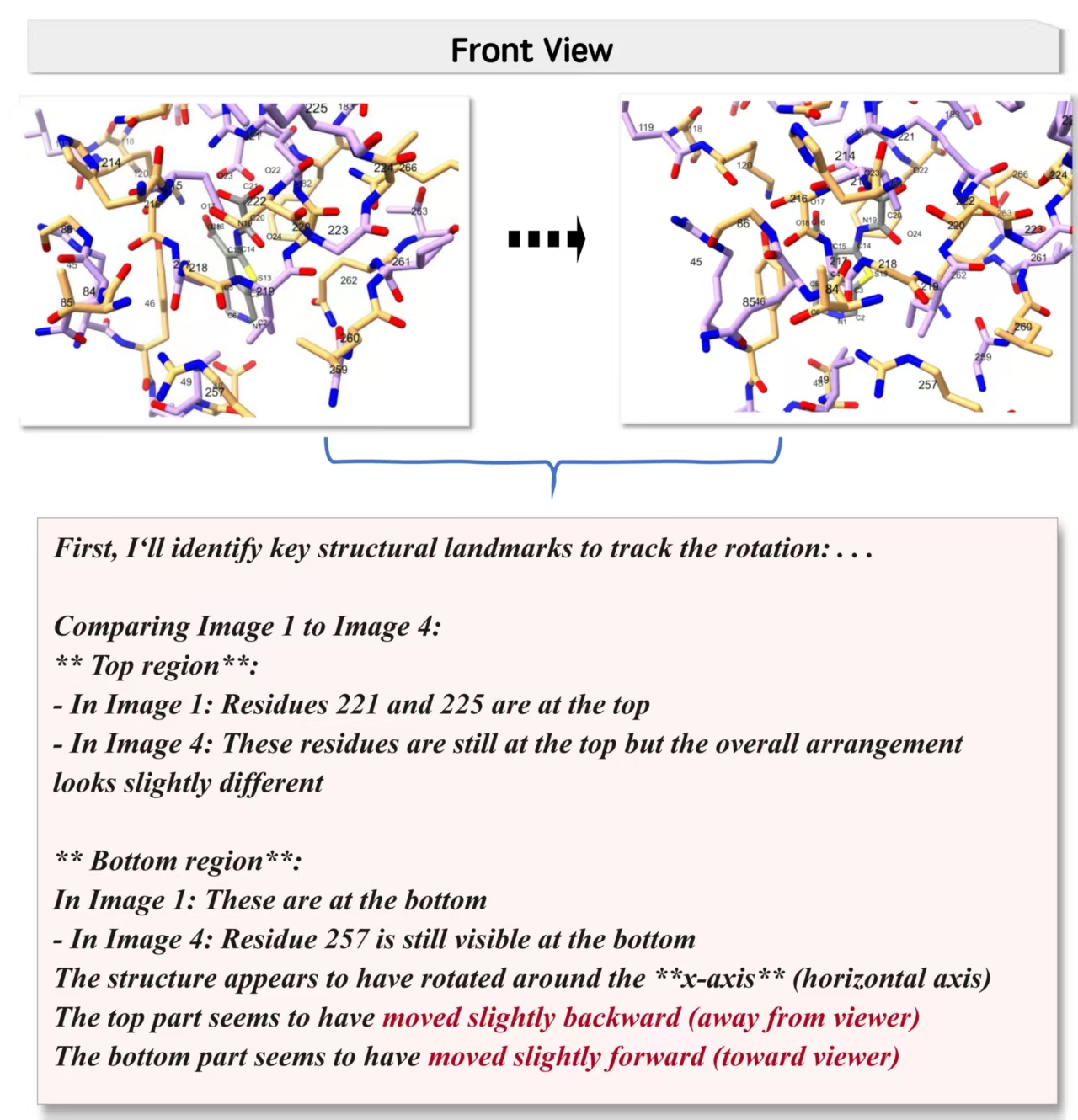}
    \caption{Case study of the Rotation task.}
    \label{fig:case}
\end{figure}

\textbf{Case Study.}
In this section, we examine Claude Sonnet 4.5, the top-performing model among advancing VLMs, by analyzing its reasoning process in failure cases from the Rotation task. As shown in ~\cref{fig:case}, the model demonstrates a logically sound approach: it identifies conserved residues across structural changes as anchors and infers the rotation axis and angle accordingly. However, in the key region it localizes, although the model correctly detects spatial rearrangements of residues 221 and 225, it misinterprets the change as "moved slightly backward," leading to an incorrect conclusion. In fact, simply observing the positional shift of residue 221 would suggest a rotation around the y-axis. This case indicates that advancing VLMs still lack adequate spatial reasoning capabilities and require more effective mechanisms to elicit such skills.

%% file: Table/Main_results.tex
\begin{table*}[]
\caption{Evaluation on \texttt{\benchmark}. Bold indicates the best result among all models. Since in the Trans-Rot. and Rot-Rot. tasks, the predictions of almost all models are close to random guessing, we exclude these two rows when calculating the average scores.}
\label{tab:main_results}
\resizebox{\textwidth}{!}{
\begin{tabular}{l|cc|cccccccccc}
    & & &
    \rotatebox{75}{Translation} &
    \rotatebox{75}{Rotation} &
    \rotatebox{75}{Zooming} & 
    \rotatebox{75}{Res-Lig Inter Pos.} &
    \rotatebox{75}{Res-Lig Inter Neg.} &
    \rotatebox{75}{Trans-Rot.} &
    \rotatebox{75}{Rot-Rot.} &
    \rotatebox{75}{Docking} &
    \rotatebox{75}{Inter Location} & 
    \rotatebox{75}{Poc-Lig Inter.} \\
    Methods & Rank & Avg. & \multicolumn{5}{c}{\cellcolor{orange!12}Unit Task} & \multicolumn{5}{c}{\cellcolor{yellow!12}Composite Task} \\
    \hline
    \addlinespace[0.2em]
    Human Level &- &81.18 &100.00 &70.18 & 30.00 & 100.00&100.00 & 32.00& 26.00& 74.54 &92.00& 82.78\\
    \rowcolor{gray!10}
    \addlinespace[0.2em]
    \multicolumn{1}{l|}{\textcolor{black}{Reasoning Models}} & & & & & & & & & & & &\\
    \addlinespace[0.2em]
    GPT-5-mini& 9 & 27.71& 47.71 & 30.55 & 4.00 & 29.33 & 34.00 & 28.00 & 22.44 & 27.24 & 47.82 & 1.01\\
    \addlinespace[0.2em]
    O4-mini& 8 & 28.55& 39.71 & 36.36 & 2.08 & 12.67 & 76.00 & 40.00 & 20.00 & 31.51 & 30.00 & 0.00\\
    \addlinespace[0.2em]
    O3& 3& 33.65& 52.29 & 43.82 & 2.00 & 18.67 & \textbf{94.00} & 22.00 & 22.44 & 20.69 & 36.00 & 1.71\\
    \addlinespace[0.2em]
    Gemini-2.5-pro& 6 & 29.94& 50.15 & 38.88 & 0.00 & 28.67 & 52.00 & 30.61 & 21.62 & 30.38 & 38.00 & 1.44\\
    \addlinespace[0.2em]
    Gemini-2.5-flash-lite& 11 & 16.00& 36.29 & 22.55 & 4.00 & 6.67 & 0.00 & 30.00 & 25.00 & 32.25 & 26.00 & 0.25\\
    \addlinespace[0.2em]
    Claude Opus4& 4 & 33.13& 57.43 & 24.73 & 6.00 & 33.67 & 74.00 & 12.00 & 26.00 & \textbf{34.39} & 34.00 & 0.77\\
    \addlinespace[0.2em]
    Claude Sonnet4.5& 2& 34.37 & 45.71 & 44.18 & 6.00 & 22.33 & 84.00 & 28.00  & 26.00 & 34.12 & 38.00 & 0.60\\
    \addlinespace[0.2em]
    Qwen3-vl-235b-a22b-thinking& 10 & 23.34& 46.36 & 25.21 & 6.00 & 17.03 & 25.00 & 20.40 & 22.00 & 29.32 & 38.00 & 0.00 \\
    \hline
    \rowcolor{gray!10}
    \addlinespace[0.2em]
    \multicolumn{1}{l|}{\textcolor{black}{General Models}} & & & & & & & & & & & &\\
    \addlinespace[0.2em]
     GPT-41& 7 & 29.20 & 29.71 & 37.45 & 2.00 & 7.33 & 80.00 & 33.33 & 29.26 & 32.90 & 36.00 & 0.2\\
    \addlinespace[0.2em]
    Claude Sonnet3.5& 5 & 31.23& 47.14 & 37.11 & 10.00 & 27.50 & 70.00 & 18.00  & 32.00 & 27.52 & 28.00 & 2.55\\
    \hline
    \rowcolor{gray!10}
    \addlinespace[0.2em]
    \multicolumn{1}{l|}{\textcolor{black}{Our Model}} & & & & & & & & & & & & \\
    Qwen2.5VL-7B-SFT & 1 & \textbf{62.96} & \textbf{99.84} & \textbf{99.71} & \textbf{27.14} & \textbf{63.46} & 89.52 & \textbf{88.44} & \textbf{89.59} & 24.94 & \textbf{88.37} & \textbf{10.72}\\ \bottomrule
    \end{tabular}
}
\end{table*}

%% file: sec/5_Conclusion.tex
\section{Conclusion}
In this work, we establish Microscopic Spatial Intelligence (MiSI) as a distinct and critical challenge for Vision-Language Models (VLMs), extending beyond macroscopic understanding to the atom-level reasoning essential for scientific discovery. We propose a systematic benchmark framework dubbed as \texttt{\benchmark}, for valuating various advanced VLMs for MiSI. The experiments reveals a significant performance gap between state-of-the-art VLMs and human expertise. Yet, the strong performance of a fine-tuned 7B model underscores the substantial potential of VLMs to master complex spatial transformations, even surpassing humans on complex spatial tasks such as rotation. Ultimately, achieving robust MiSI will require not only scaling model architectures but also the explicit integration of scientific knowledge for real-world scientific applications.

%% file: sec/suppl_A.tex
\clearpage
\setcounter{page}{1}

\begin{figure*}[]
    \centering    \includegraphics[width=1.0\linewidth]{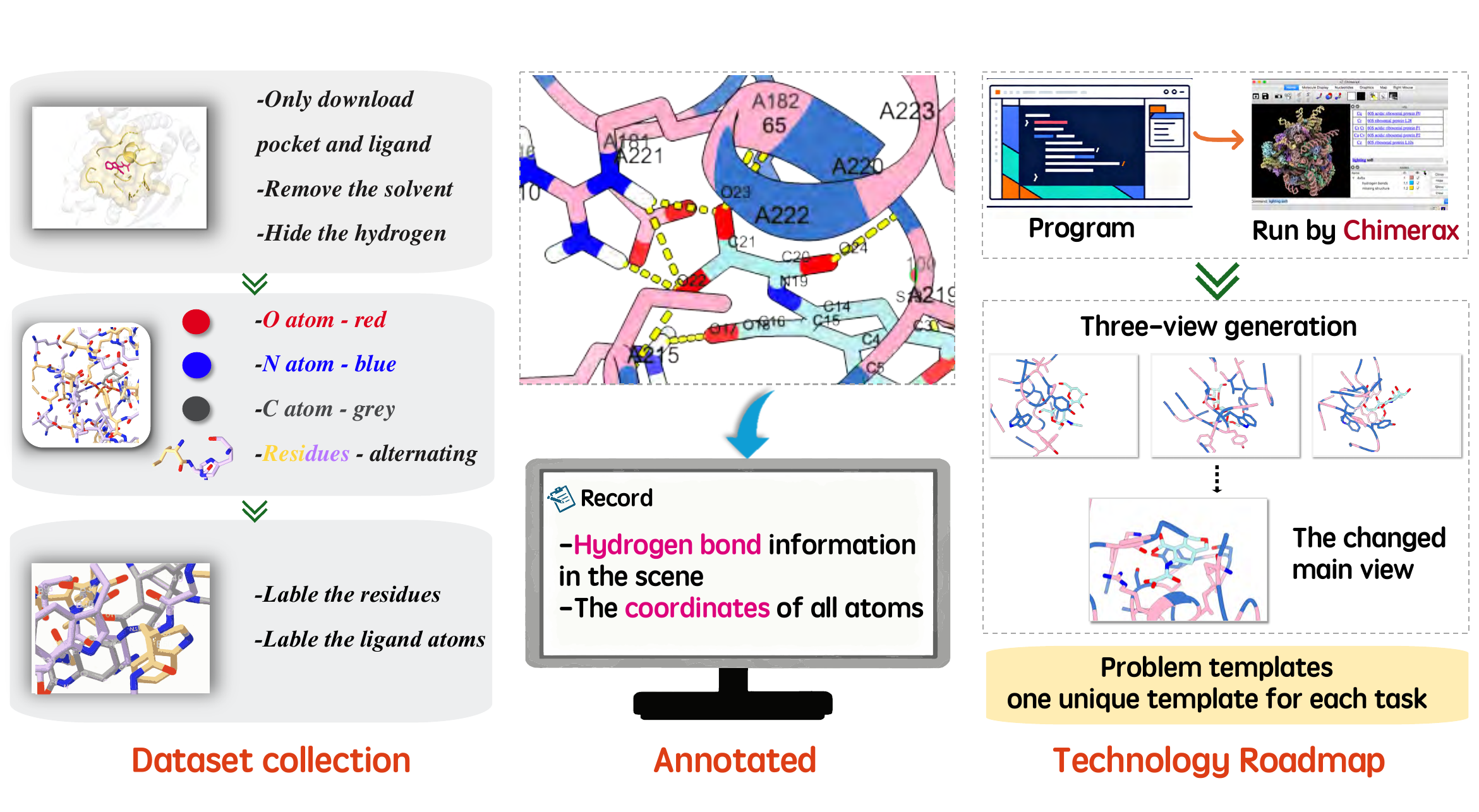}
    \caption{Data generation pipeline. Our pipeline comprises three key stages: data collection and filtering, data annotation, and a unified module for synthesizing both images and question-answer pairs using Chimera with fixed templates.}
    \label{fig:pipeline}
    \vspace{-15pt}
\end{figure*}

\section{Dataset Details}
In this section, we present additional details regarding the construction of the dataset.
\subsection{Data Generation Pipeline}



As shown in ~\cref{fig:pipeline}, our dataset construction process consists of three main steps.

The first step is Dataset Collection. We download the PDBBind dataset~\citep{wang2005pdbbind}, retaining only the ligand and pocket (collectively referred to as the complex) for each sample. 
As the first exploratory dataset for microscopic spatial intelligence, we reduce the overall complexity by removing the solvent and hiding the hydrogen. Subsequently, we color all oxygen, nitrogen, and carbon atoms red, blue, and gray, respectively, in accordance with common coloring standards in the field. Notably, for pocket residues, we apply alternating yellow and purple coloring to facilitate the model's ability to distinguish adjacent residues in subsequent interaction (hydrogen bond) recognition tasks.

The second step is Annotation. For each complex, we use the ChimeraX~\citep{pettersen2021ucsf} command \texttt{get sel screen} to obtain the screen coordinates of all atoms. We then employ ChimeraX's built-in hydrogen bond calculation function to identify all hydrogen bonds between the pocket and ligand. All this information is annotated for every complex and stored for direct use in constructing subsequent subtask-specific datasets.

The third step is Subtask-Specific Data Generation. For each subtask, we first develop corresponding code according to its requirements. We then use ChimeraX to generate multiple image samples for every complex in the training and test sets. Finally, we design question-answering (QA) templates for each subtask and populate them with the meta-information of each sample to form complete data instances.

\subsection{Zooming}

\begin{figure}[h!]
    \centering
    \includegraphics[width=\columnwidth]{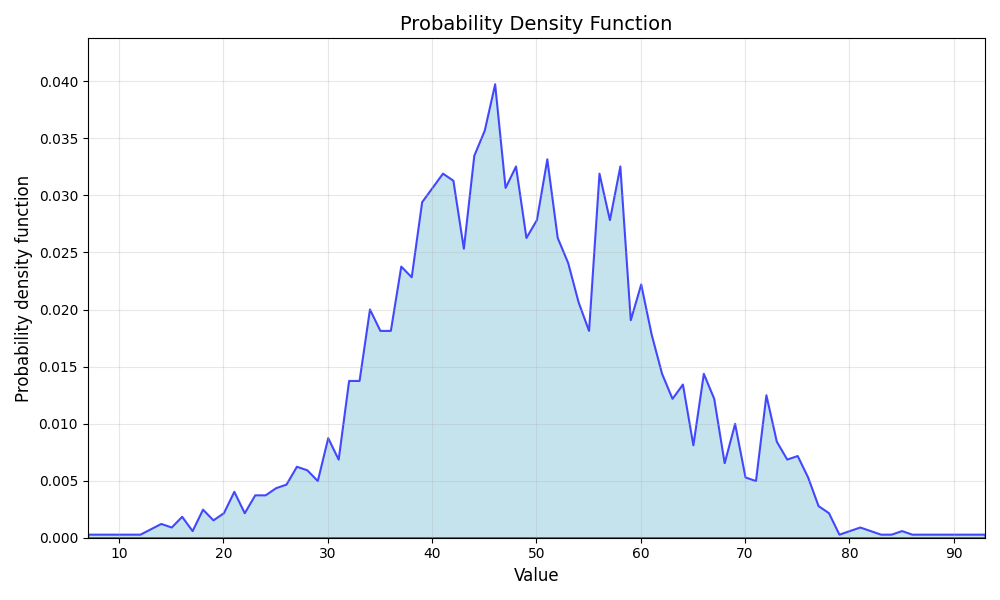}
    \caption{The probability density function of the movement depth.}
    \label{fig:z_pdf}
    \vspace{-15pt}
\end{figure}


\begin{figure}[h!]
    \centering
    \includegraphics[width=\columnwidth]{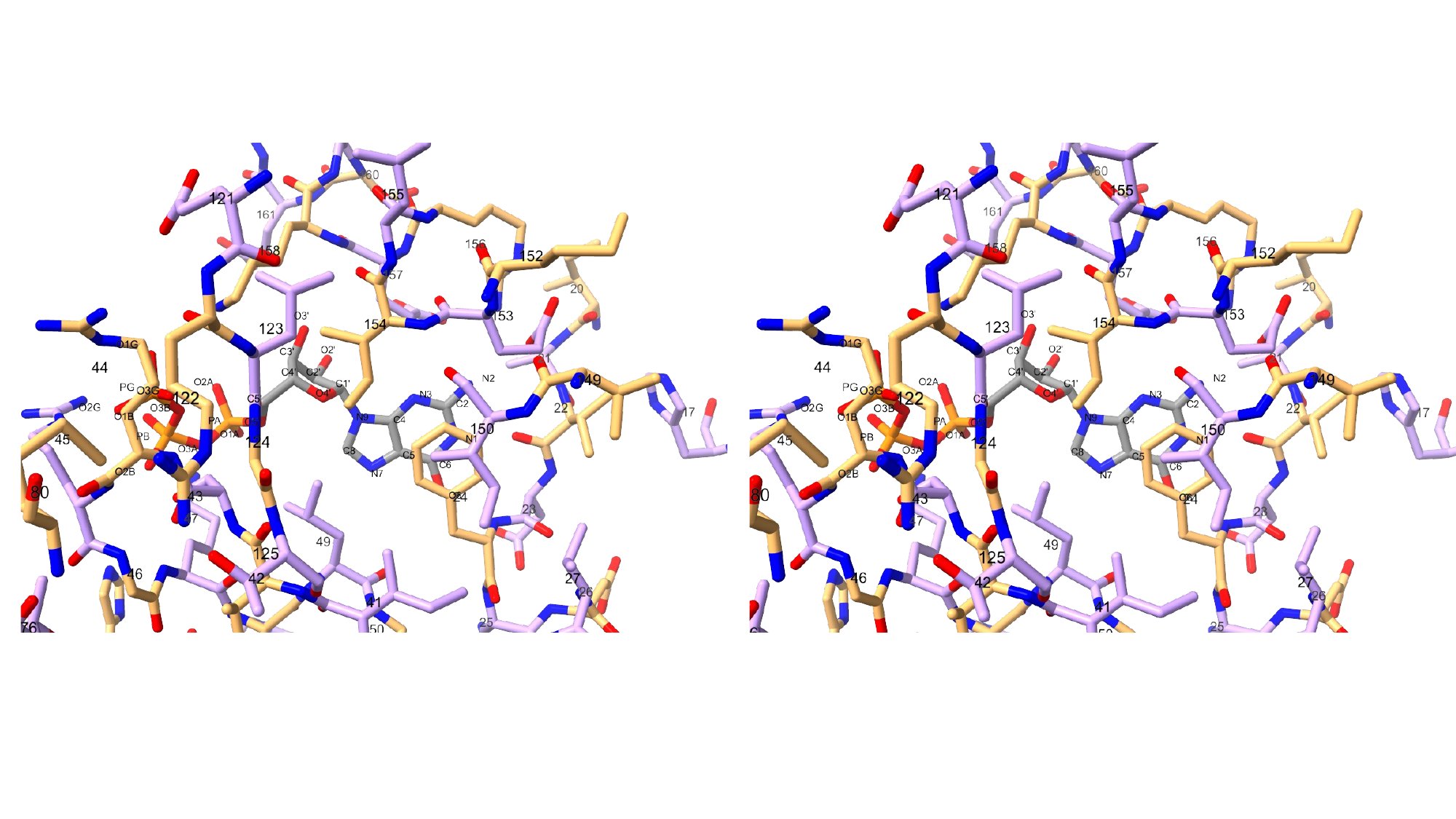}
    \caption{Comparison of different movement depths.}
    \label{fig:move_z_compare}
    \vspace{-15pt}
\end{figure}

For the zooming task, we compute the probability density function of the movement depth required to translate all interactions from their initial states to the screen center in the training set, as shown in ~\cref{fig:z_pdf}. The results indicate that most movement depths fall within the range of 20–80 angstroms (\AA)). Through empirical testing, we observe that when the movement depth lies in the 20–40 \AA interval, a one-unit difference in the zooming operation (e.g., \texttt{move z 25} vs. \texttt{move z 26}) produces only minor changes in the output, making them difficult to distinguish and thus unsuitable as training data (as illustrated in ~\cref{fig:move_z_compare}). On the other hand, for movement depths in the 60–80 \AA range, due to the varying spatial conformations of different PDB structures in their initial states, some structures become zoomed in to the extent that only a single atom remains visible when depth values exceed 60 \AA. Such samples are likewise inadequate for training the model to discern specific zooming depth values. Therefore, we exclusively select depth values within the 40–60 \AA interval to generate our dataset.

\subsection{Residue-Ligand Interaction}

In tasks related to interaction, the calculation of ground-truth hydrogen bonds follows the default protocol of the software ChimeraX~\citep{pettersen2021ucsf}. The distance and angle cutoffs for hydrogen bonding are based on a survey of small-molecule crystal structures, as described in ChimeraX.
Additionally, the option to relax distance and angle criteria—that is, whether to incorporate tolerance values beyond the precise criteria for identifying hydrogen bonds (which involve several distinct distance and angle thresholds depending on the atom types involved)—is also adopted from the reference provided in ChimeraX. Specifically, a distance tolerance of 0.4 \AA and an angle tolerance of 20 degrees are used.

\subsection{Ligand Docking }


In ligand docking task, to enable the model to better observe the respective conformational and geometric information of the protein pocket and the displaced ligand, it is essential to minimize the overlapping region between the displaced ligand and the pocket. In ChimeraX, most molecular complexes initially occupy the full vertical extent (e.g., the Y-axis) of the screen; therefore, we only consider translating the ligand horizontally (e.g., along the X-axis) to either the far left or far right side of the screen. We randomly select the first complex (PDB ID: 1ugx) from the training dataset as a reference. In its native docking conformation, the mean screen coordinates of all atoms in this complex are denoted as $(X^{c}_{base}, Y^{c}_{base}, Z^{c}_{base})$. When the ligand alone is moved to the far right side of the screen, 
the mean screen coordinates of all atoms in the ligand become $(X^{l}_{base}, Y^{l}_{base}, Z^{l}_{base})$.

For other complexes in the dataset, the mean screen coordinates of all atoms in each complex under the native docking conformation are obtained as $(x^{c}, y^{c}, z^{c})$. Similarly, under the same conformation, the maximum and minimum X-axis screen coordinates of all atoms in the ligand are denoted as $(x^{l}_{max}, x^{l}_{min})$. Based on this, the distance required to move to the farthest point is approximated according to the Field of View. Specifically, the distance to move to the far-right is calculated as $dst_{r} = -z^{c} * X^{l}_{base} /Z^{c}_{base} - x^{l}_{max}$, and the distance to move to the far-left is $dst_{l} = -(z^{c} * X^{l}_{base} / Z^{c}_{base} - x^{l}_{max})$. Subsequently, numerical values are randomly sampled from $[0, 1, 2]$ and subtracted from or added to $dst_r$ and $dst_l$, respectively. These adjusted distances are then combined with subsequent rotation operations to generate additional dataset samples.

%% file: sec/suppl_B.tex
\section{More Results}
\input{Table/main_results_ema}



In this section, we present the Exact Matching Accuracy of all evaluated models and human evaluators on the \texttt{\benchmark} dataset. A prediction is considered correct only if it exactly matches the ground-truth. The corresponding results are shown in Table~\cref{tab:main_results_ema}. It should be noted that for the Protein–Ligand Interaction task, while the original test set contains a certain number of complexes without hydrogen bonds, there is only one such complex in \texttt{\benchmark (tiny)}. Given that predicting the absence of hydrogen bonds is considerably simpler than predicting all hydrogen bonds correctly, we report the Exact Matching Accuracy of the models only for complexes that contain hydrogen bonds.

From the table, we can observe that when Exact Matching Accuracy is applied, the performance gap between all advancing VLMs and our SFT model becomes more pronounced. Particularly in the Translation and Rotation tasks, the transformation of the metric have a limited impact on the SFT model, whereas the performance of advancing VLMs exhibits a substantial decline. This indicates that while advancing VLMs possess strong potential for spatial understanding and reasoning, effective methods are required to activate this capability. For the SFT model, the performance gap with human evaluators is further widened in tasks related to interaction recgnition. This indicates that current models lack specialized domain knowledge, and suggests that incorporating such knowledge during the pre-training phase may be necessary for progressing toward more general artificial intelligence.

%% file: Table/main_results_ema.tex
\begin{table*}[t]
\caption{Evaluation on \texttt{\benchmark}. We employ Exact Matching Accuracy as the evaluation metric. Bold indicates the best result among all models. Since in the Trans-Rot. and Rot-Rot. tasks, the predictions of almost all models are close to random guessing, we exclude these two rows when calculating the average scores.}
\label{tab:main_results_ema}
\resizebox{\textwidth}{!}{
\begin{tabular}{l|cc|cccccccccc}
    & & &
    \rotatebox{75}{Translation} &
    \rotatebox{75}{Rotation} &
    \rotatebox{75}{Zooming} & 
    \rotatebox{75}{Res-Lig Inter Pos.} &
    \rotatebox{75}{Res-Lig Inter Neg.} &
    \rotatebox{75}{Trans-Rot.} &
    \rotatebox{75}{Rot-Rot.} &
    \rotatebox{75}{Docking} &
    \rotatebox{75}{Inter Location} & 
    \rotatebox{75}{Poc-Lig Inter.} \\
    Methods & Rank & Avg. & \multicolumn{5}{c}{\cellcolor{orange!12}Unit Task} & \multicolumn{5}{c}{\cellcolor{yellow!12}Composite Task} \\
    \hline
    \addlinespace[0.2em]
    Human Level &- &76.25 &100.00 &58.00 & 30.00 & 100.00&100.00 & 32.00& 26.00& 60.00 &92.00& 70.00\\
    \hline
    \rowcolor{gray!10}
    \addlinespace[0.2em]
    \multicolumn{1}{l|}{\textcolor{black}{Reasoning Models}} & & & & & & & & & & & &\\
    \addlinespace[0.2em]
    GPT-5-mini& 8 & 16.23& 10.00 & 10.00 & 4.00 & 22.00 & 34.00 & 28.00 & 22.44 & 2.04 & 47.82 & 0.00 \\
    \addlinespace[0.2em]
    O4-mini& 9 & 13.01& 6.00 & 12.00 & 2.08 & 8.00 & 76.00 & 40.00 & 20.00 & 0.00 & 30.00 & 0.00\\
    \addlinespace[0.2em]
    O3& 2& 33.65& 22.00 & 20.00 & 2.00 & 12.00 & \textbf{94.00} & 22.00 & 22.44 & 0.00 & 36.00 & 0.00\\
    \addlinespace[0.2em]
    Gemini-2.5-pro& 6 & 17.54& 14.89 & 23.40 & 0.00 & 12.00 & 52.00 & 30.61 & 21.62 & 0.00 & 38.00 & 0.00\\
    \addlinespace[0.2em]
    Gemini-2.5-flash-lite& 11 & 9.04& 36.29 & 2.00 & 4.00 & 4.00 & 0.00 & 30.00 & 25.00 & 0.00 & 26.00 & 0.00\\
    \addlinespace[0.2em]
    Claude Opus4& 4 & 19.54 & 14.29 & 6.00 & 6.00 & 22.00 & 74.00 & 12.00 & 26.00 & 0.00 & 34.00 & 0.00\\
    \addlinespace[0.2em]
    Claude Sonnet4.5& 3& 20.75 & 8.00 & 14.00 & 6.00 & 14.00 & 84.00 & 28.00  & 26.00 & 2.00 & 38.00 & 0.00\\
    \addlinespace[0.2em]
    Qwen3-vl-235b-a22b-thinking& 10 & 11.80& 14.29 & 4.55 & 6.00 & 6.52 & 25.00 & 20.40 & 22.00 & 0.00 & 38.00 & 0.00 \\
    \hline
    \rowcolor{gray!10}
    \addlinespace[0.2em]
    \multicolumn{1}{l|}{\textcolor{black}{General Models}} & & & & & & & & & & & &\\
    \addlinespace[0.2em]
     GPT-41& 7 & 17.50 & 4.00 & 16.00 & 2.00 & 2.00 & 80.00 & 33.33 & 29.26 & 0.00 & 36.00 & 0.00\\
    \addlinespace[0.2em]
    Claude Sonnet3.5& 5 & 18.02& 10.00 & 8.16& 10.00 & 18.00 & 70.00 & 18.00 & 32.00 & 0.00 & 28.00 & 0.00\\
    \hline
    \rowcolor{gray!10}
    \addlinespace[0.2em]
    \multicolumn{1}{l|}{\textcolor{black}{Our Model}} & & & & & & & & & & & & \\
    Qwen2.5VL-7B-SFT & 1 & \textbf{57.56} & \textbf{98.88} & \textbf{97.48} & \textbf{27.14} & \textbf{57.52} & 89.52 & \textbf{88.44} & \textbf{89.59} & 0.75 & \textbf{88.37} & \textbf{0.78}\\ \bottomrule
    \end{tabular}
}
\end{table*}

%% file: sec/suppl_C.tex
\section{Visualization}

\begin{figure*}[]
    \centering    \includegraphics[width=0.65\linewidth]{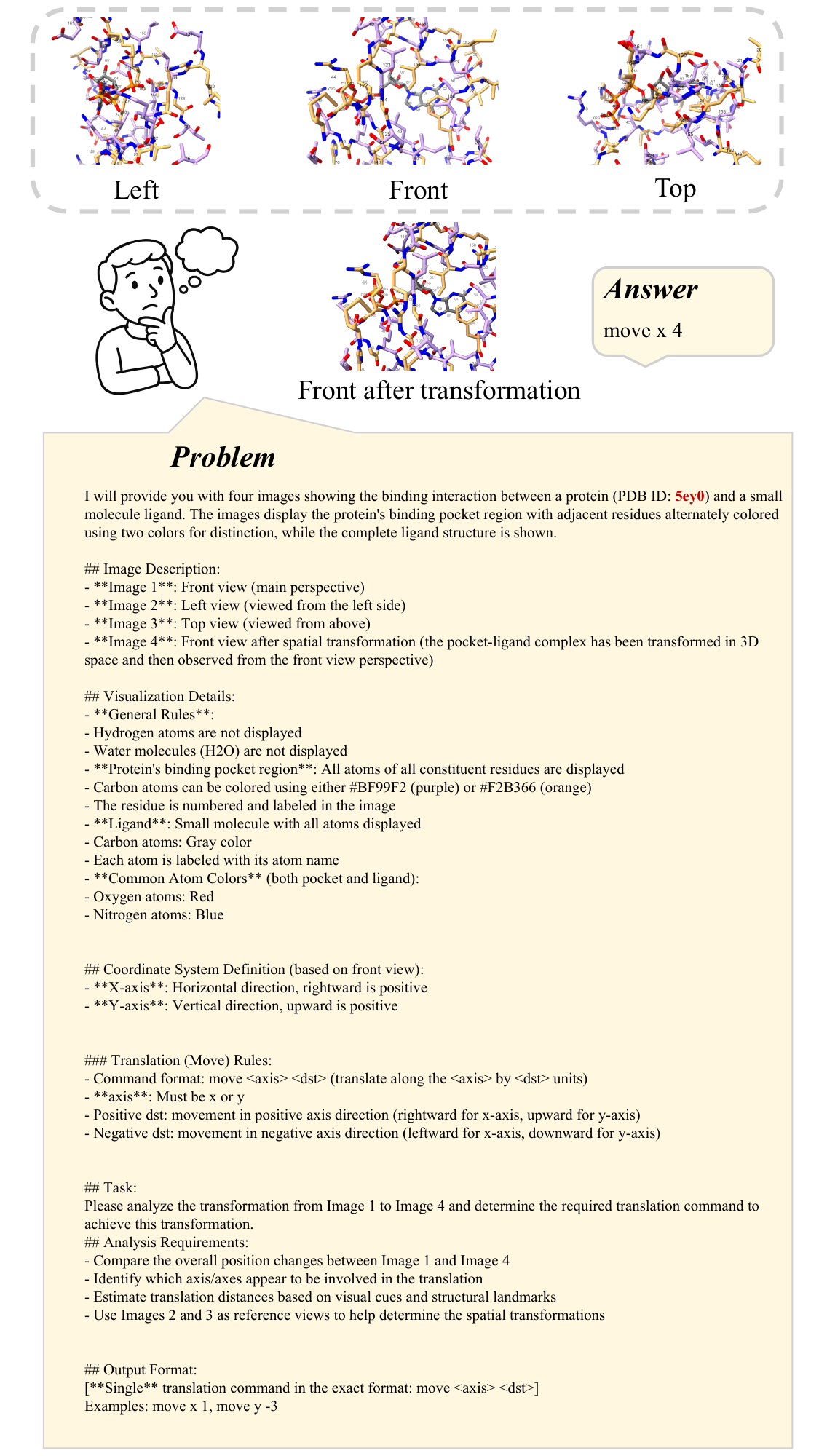}
    \caption{Sample visualization for the translation task. Zoom in for greater detail.}
    \label{fig:translation}
\end{figure*}
\begin{figure*}[]
    \centering    \includegraphics[width=0.65\linewidth]{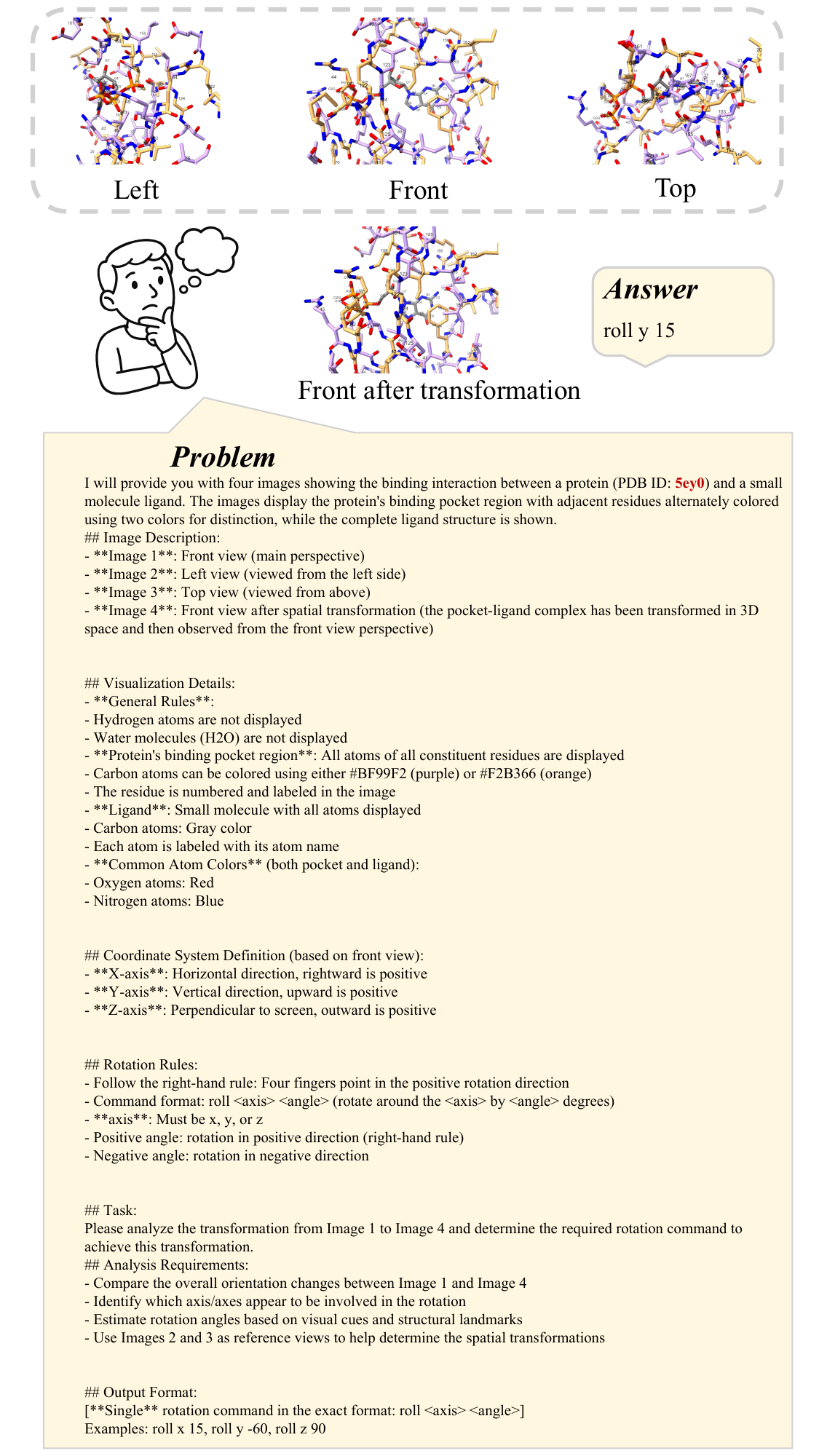}
    \caption{Sample visualization for the rotation task. Zoom in for greater detail.}
    \label{fig:rotation}
\end{figure*}
\begin{figure*}[]
    \centering    \includegraphics[width=0.65\linewidth]{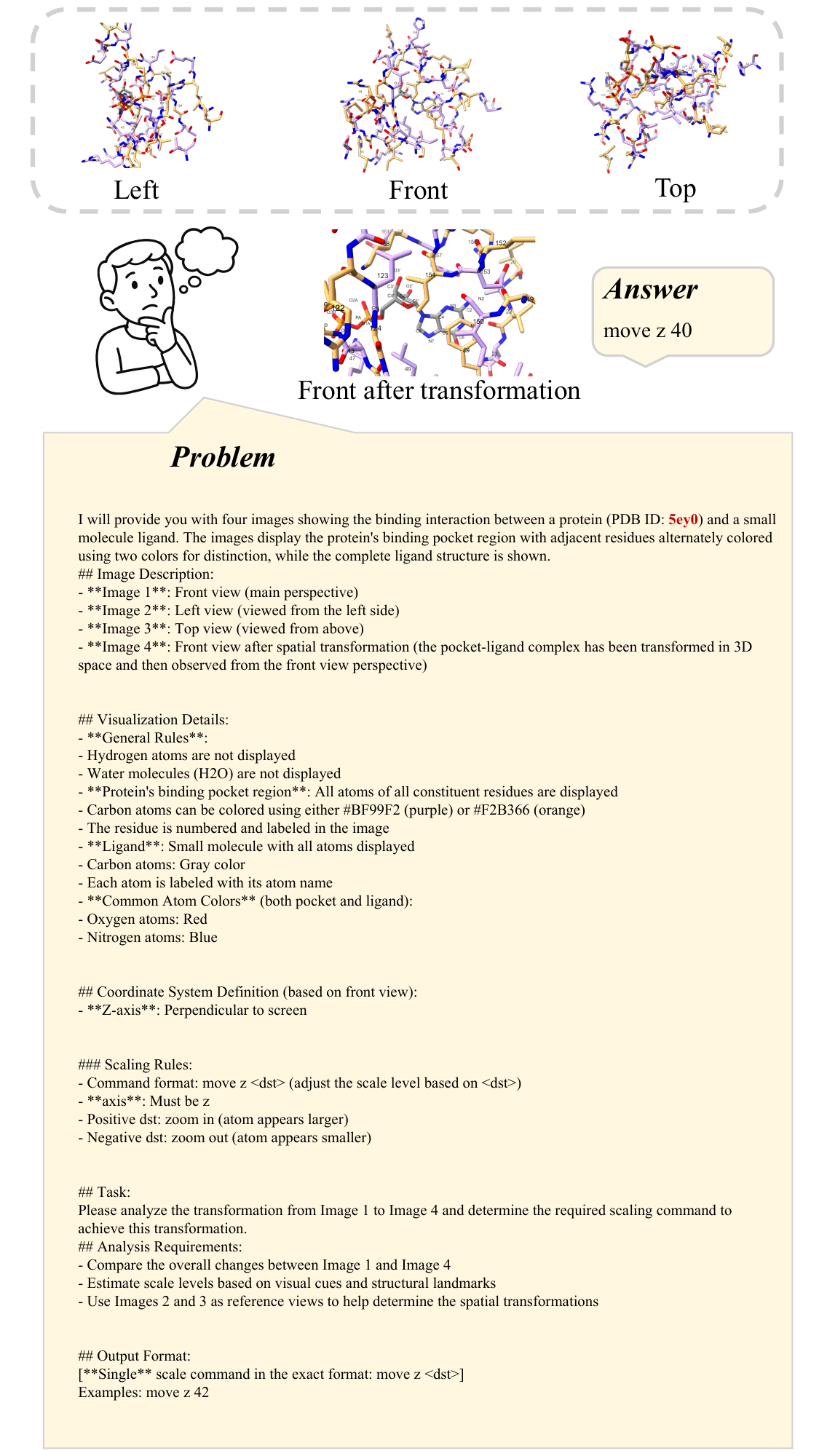}
    \caption{Sample visualization for the zooming task. Zoom in for greater detail.}
    \label{fig:zooming}
\end{figure*}
\begin{figure*}[]
    \centering    \includegraphics[width=0.65\linewidth]{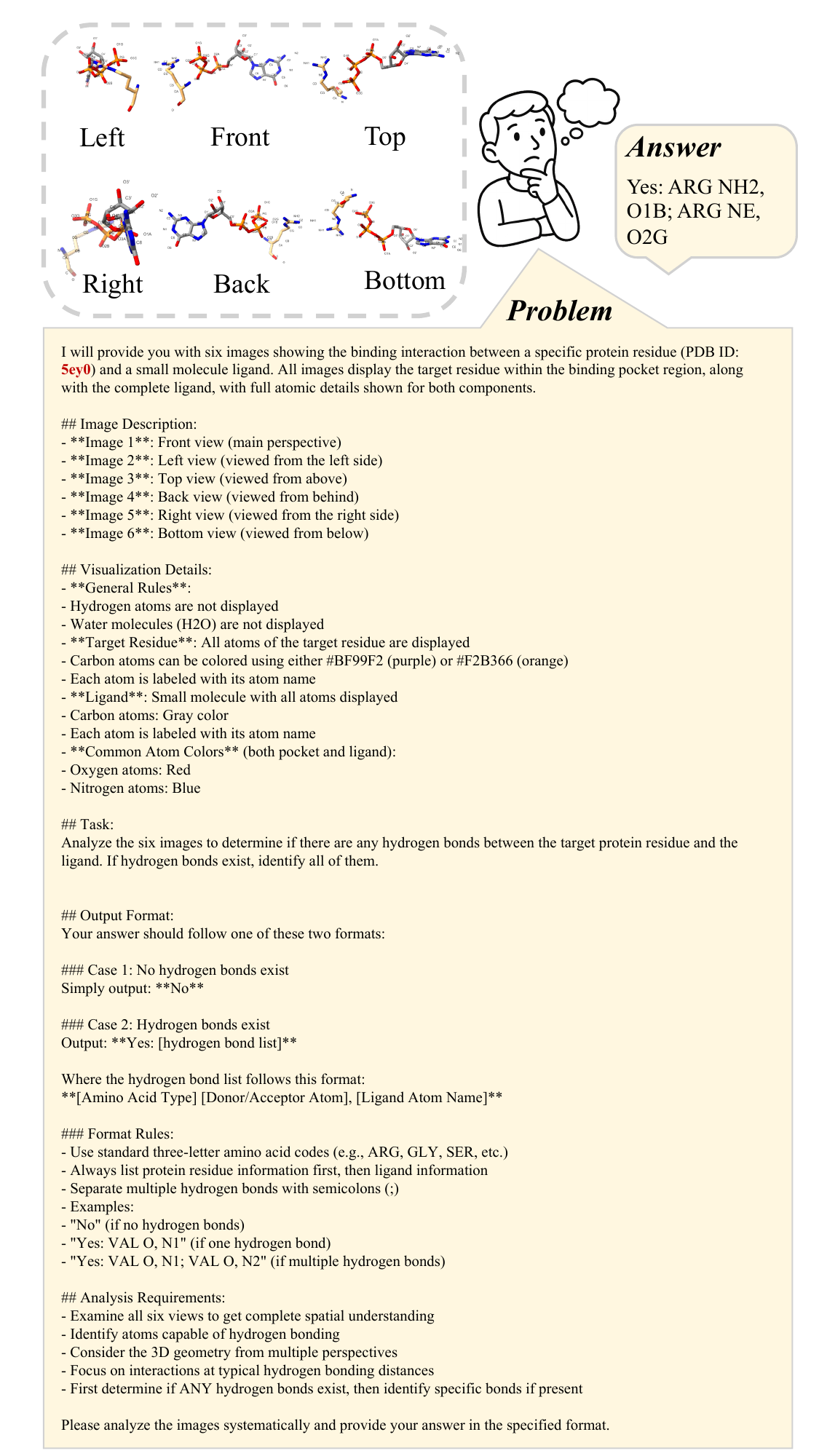}
    \caption{Sample visualization for the residue-ligand interaction task. Zoom in for greater detail.}
    \label{fig:residue-ligand}
\end{figure*}
\begin{figure*}[]
    \centering    \includegraphics[width=0.65\linewidth]{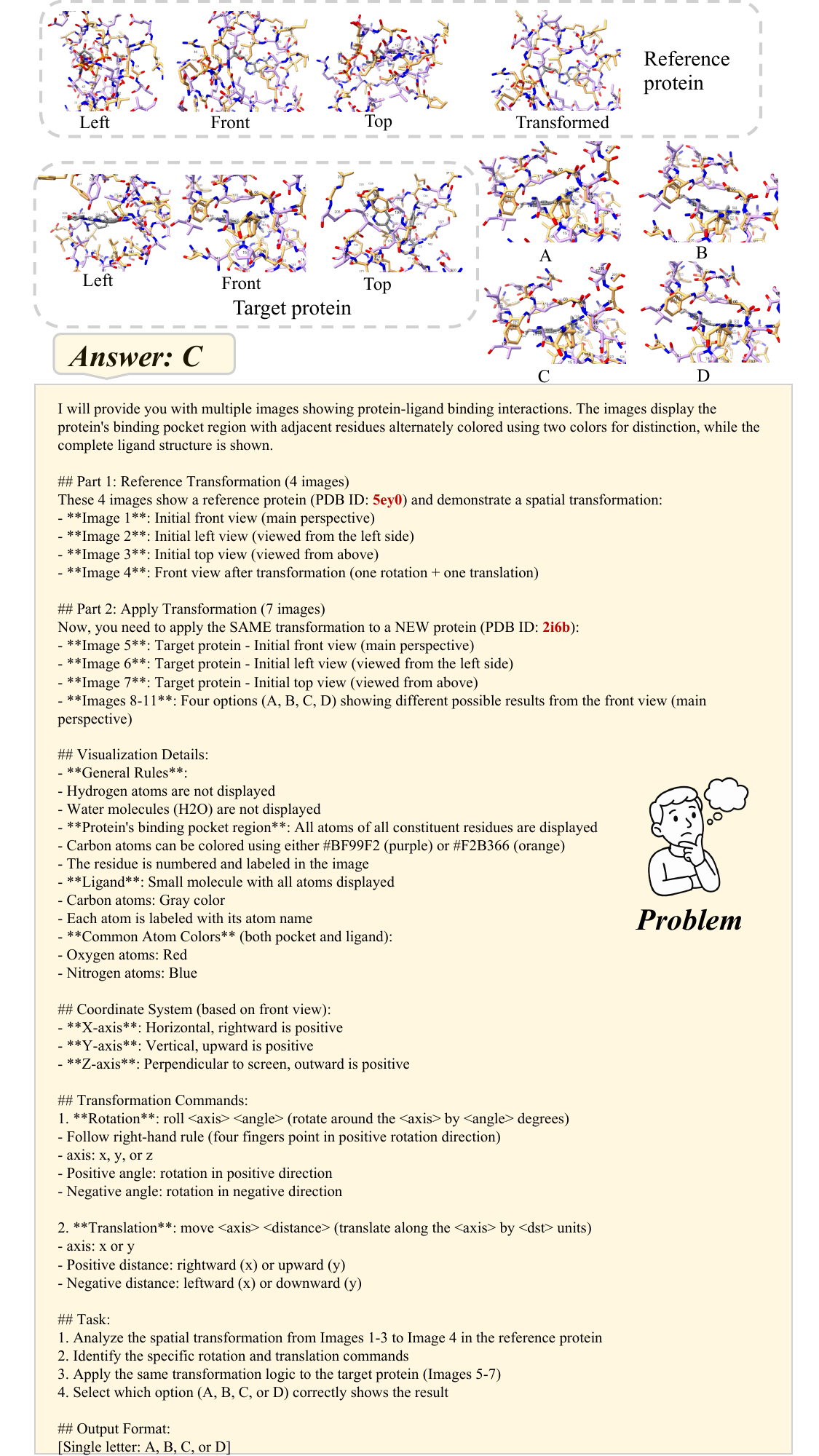}
    \caption{Sample visualization for the translation-rotation movement task. Zoom in for greater detail.}
    \label{fig:translation-rotation}
\end{figure*}
\begin{figure*}[]
    \centering    \includegraphics[width=0.65\linewidth]{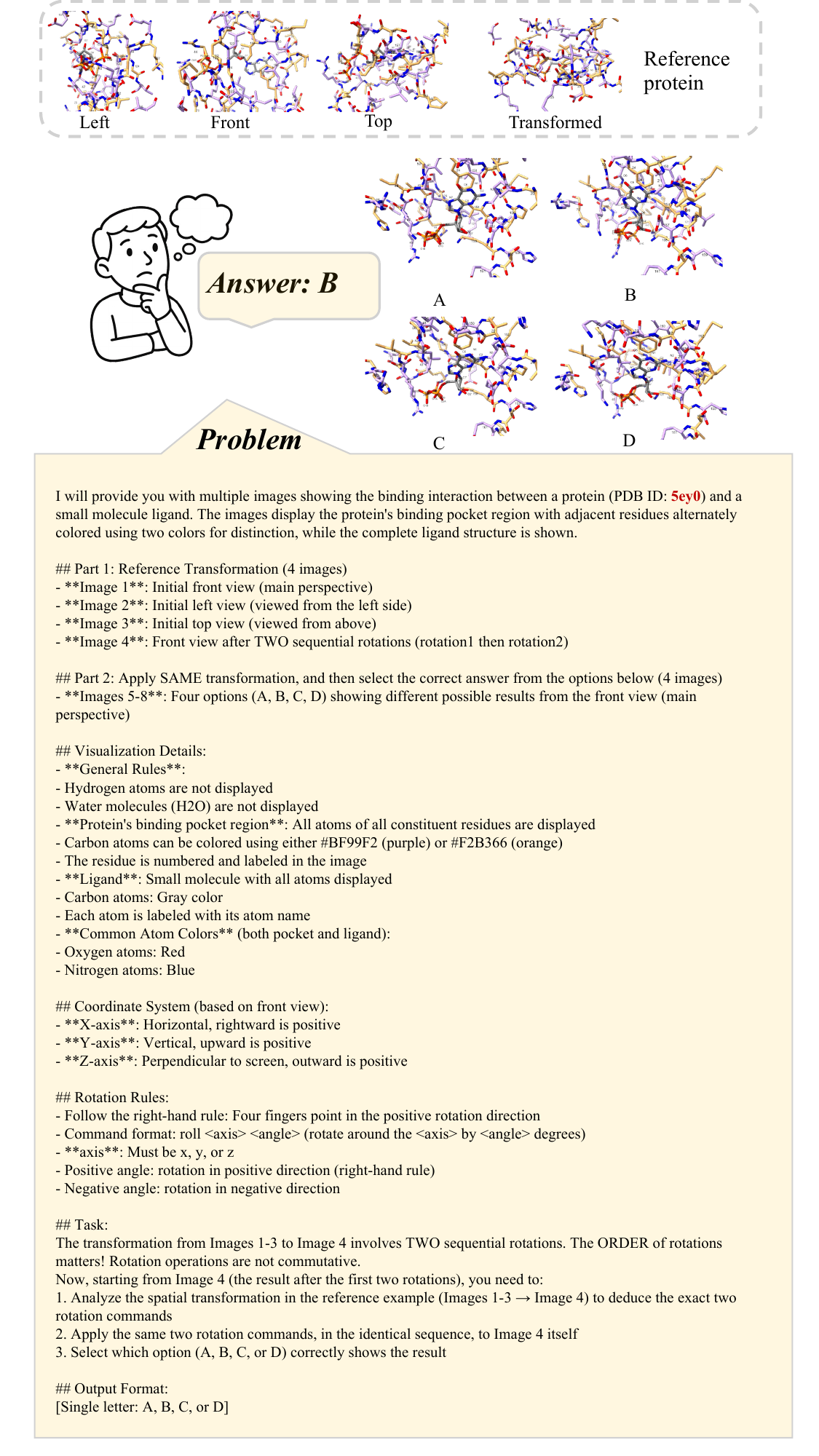}
    \caption{Sample visualization for the rotation-rotation movement task. Zoom in for greater detail.}
    \label{fig:rotation-rotation}
\end{figure*}
\begin{figure*}[]
    \centering    \includegraphics[width=0.65\linewidth]{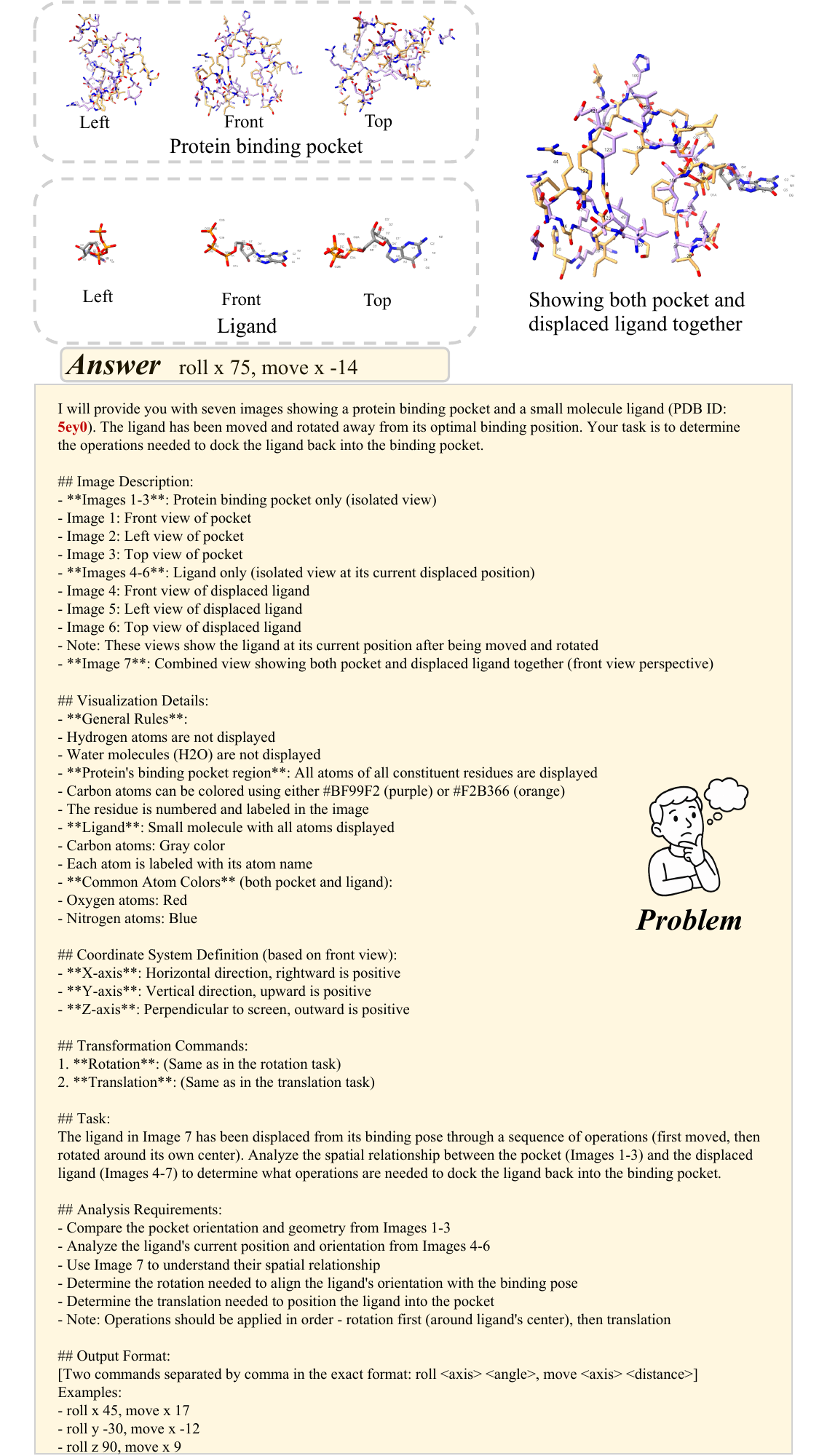}
    \caption{Sample visualization for the ligand docking task. Zoom in for greater detail.}
    \label{fig:docking}
\end{figure*}
\begin{figure*}[]
    \centering    \includegraphics[width=0.65\linewidth]{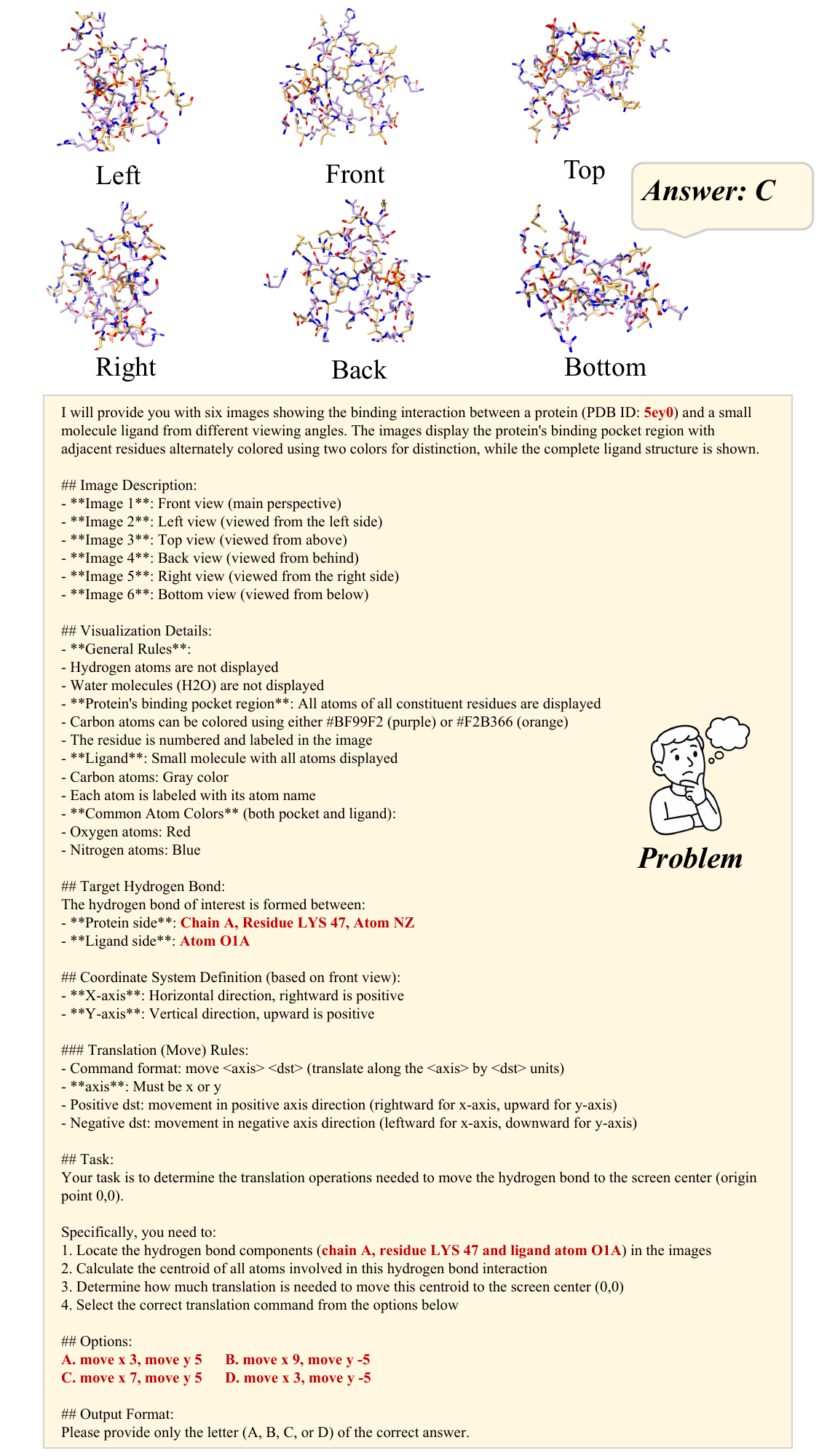}
    \caption{Sample visualization for the interaction location task. Zoom in for greater detail.}
    \label{fig:location}
\end{figure*}
\begin{figure*}[]
    \centering    \includegraphics[width=0.65\linewidth]{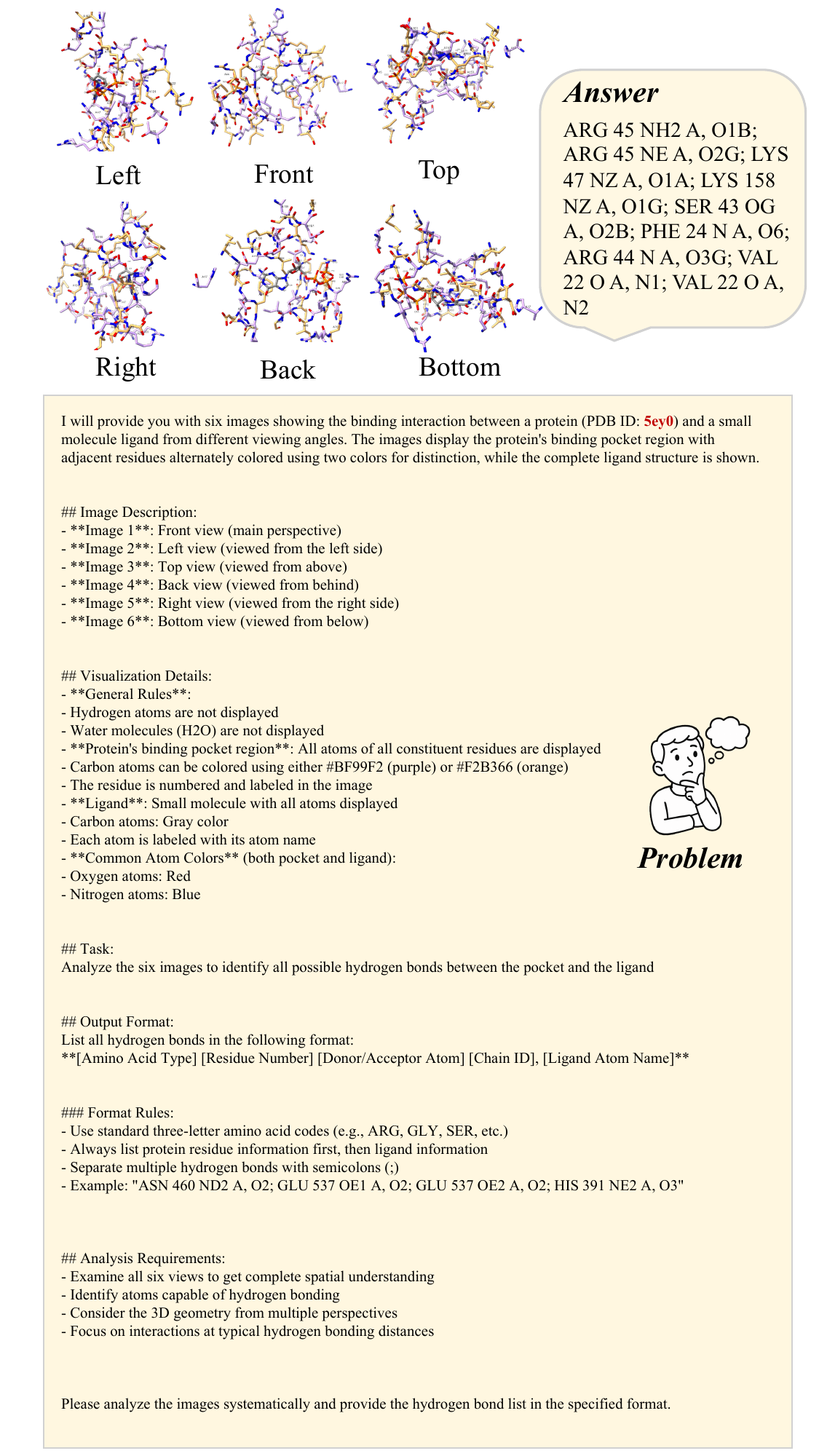}
    \caption{Sample visualization for the pocket-ligand interaction task. Zoom in for greater detail.}
    \label{fig:pocket-ligand}
\end{figure*}

In this section, we present complete examples for all nine tasks, as described in ~\cref{fig:translation,fig:rotation,fig:zooming,fig:residue-ligand,fig:translation-rotation,fig:rotation-rotation,fig:docking,fig:location,fig:pocket-ligand}, respectively. For display purposes, the background color of the images has been changed from black to transparent. The highlighted content in each question represents unique information for that specific sample, while the remaining text in black constitutes the unified prompt for the subtask.

